\documentclass{article}

\usepackage{microtype}
\usepackage{graphicx}
\usepackage{soul}
\usepackage{subfigure}
\usepackage{booktabs,xcolor} % for professional tables
\usepackage{multirow}

\usepackage{hyperref}

\usepackage[accepted]{icml2023}

\usepackage{amsmath}
\usepackage{amssymb}
\usepackage{mathtools}
\usepackage{amsthm}

\DeclareMathOperator*{\argmax}{argmax}
\DeclareMathOperator*{\topm}{top\mathnormal{-m}}

\usepackage[capitalize,noabbrev]{cleveref}

\theoremstyle{plain}

\theoremstyle{definition}

\theoremstyle{remark}

\usepackage[textsize=tiny]{todonotes}

\begin{document}

\twocolumn[
\icmltitle{Learning to Design Analog  Circuits to Meet Threshold Specifications}



\icmlsetsymbol{equal}{*}

\begin{icmlauthorlist}
\icmlauthor{Dmitrii Krylov}{yyy}
\icmlauthor{Pooya Khajeh}{yyy}
\icmlauthor{Junhan Ouyang}{yyy}
\icmlauthor{Thomas Reeves}{yyy}
\icmlauthor{Tongkai Liu}{yyy}
\icmlauthor{Hiba Ajmal}{yyy}
\icmlauthor{Hamidreza Aghasi}{yyy}
\icmlauthor{Roy Fox}{yyy}
\end{icmlauthorlist}

\icmlaffiliation{yyy}{University of California, Irvine}

\icmlcorrespondingauthor{Dmitrii Krylov}{dkrylov@uci.edu}
\icmlcorrespondingauthor{Hamidreza Aghasi}{haghasi@uci.edu}
\icmlcorrespondingauthor{Roy Fox}{royf@uci.edu}

\icmlkeywords{Machine Learning, ICML}

\vskip 0.3in
]



\printAffiliationsAndNotice{}  % leave blank if no need to mention equal contribution

\begin{abstract}
Automated design of analog and radio-frequency circuits using supervised or reinforcement learning from simulation data has recently been studied as an alternative to manual expert design. It is straightforward for a design agent to learn an inverse function from desired performance metrics to circuit parameters. However, it is more common for a user to have \emph{threshold} performance criteria rather than an exact target vector of feasible performance measures. In this work, we propose a method for generating from simulation data a dataset on which a system can be trained via supervised learning to design circuits to meet threshold specifications. We moreover perform the to-date most extensive evaluation of automated analog circuit design, including experimenting in a significantly more diverse set of circuits than in prior work, covering linear, nonlinear, and autonomous circuit configurations, and show that our method consistently reaches success rate better than 90\% at 5\% error margin, while also improving data efficiency by upward of an order of magnitude. A demo of this system is available at \href{circuits.streamlit.app}{circuits.streamlit.app}
\end{abstract}

\section{Introduction}

% \begin{itemize}
%     \item Analog circuit design is valuable because..., but it is currently done manually, which is a burden because...
%     \item It was recently proposed to do this with deep RL agent, in a limited variety of topologies, and with a lot of simulation runs
%     \item We propose to do this with supervised learning
%     \item We envision an interface where a user specifies thresholds, we propose a method for constructing the dataset for this
%     \item Outline of experiments, results, conclusions
% \end{itemize}

Owing to the immense growth of consumer electronics over the last few decades, integrated circuitry using commercial CMOS/BiCMOS chip technologies has become a major sector of the semiconductor industry \cite{kamal}.
% \rc{In line with the Chips and Science ACT that was inducted by the US governement in 2022, artificial intelligence should be incorporated to foster next-generation of ``smart circuit design'' to realize non-existing systems due to the human design limitations.} %Following the invention of MOS transistors in 1959 \cite{MOS}, rapid development of circuits with miniaturized dimensions gained a critical momentum and exponential development, commonly evaluated by “Moore's law” \cite{1997moore}. %This rapidly increased density of transistors on CMOS circuits has enabled a large group of applications that heavily rely on analog and digital circuits. 
More specifically, fast innovation and skyrocketing demand in several industry segments, such as wireless communication and high-resolution imaging systems, has been driving interest in analog, radio-frequency, and millimeter-wave circuits and systems \cite{kamal}. %\st{and analog circuits into focus %of many industrial and academic groups 
%and it is currently one of the major focuses of the integrated circuit design industry~}.
Despite the economic and technological importance of these types of circuits, %the component of human-to-machine interaction to improve the performance and reduce the design duration, has not been matured proportional to the advances in technologies and the metrics of the circuits. Still in 2022, many research groups (us included) and 
contemporary design in research and industry  is still predominantly manual, using advanced electronic design automation tools such as the Cadence Virtuoso \cite{cadence} and the Keysight ADS~\cite{keysight} circuit simulators.
This heavy reliance on human design slows down and raises the costs of the development of future generations of electronic systems and should inevitably shift toward a more interactive design approach where humans and machines co-design analog circuits substantially faster.

A recent growing literature on automated circuit design has considered the problem of finding the parameters of components in a given circuit that would induce a desired set of performance metrics \cite{mina2022}.
Learning to output such circuit parameters is typically framed in the supervised learning setting, where a model in a given model class — often a neural network — is trained on a dataset of simulated parameter–metrics pairs to solve the \emph{inverse problem} of mapping target performance metrics to circuit parameters that meet these requirements.
%It has recently been proposed to address the same problem with reinforcement learning, where an agent is trained to perturb the inputs to a circuit simulator to meet observed performance metrics.
%In this approach, a dataset of parameters–metrics pairs is first collected by simulating the circuit a large number of times, each with different circuit parameters, and measuring the resulting performance in each simulation.
%Then a model is trained on the \emph{inverse problem} by feeding the metrics as inputs and predicting circuit parameters that minimize a loss from the ground-truth parameters that induced the input metrics.
%The simulator is executed after each perturbation and the agent obtains a reward shaped to encourage having the performance measured by the simulator meet the performance targets.
A limiting factor in this approach is the large number of times that the circuit needs to be simulated to collect enough data for accurate learning. %, both during training and during execution\rf{is this true?}.
As we aim to support larger and more intricate circuits, precise simulation becomes slow, and data efficiency requisite.

%We propose a supervised learning method that directly solves this \emph{inverse problem} by regression.
%Given a simulator that takes in circuit parameters and measures the circuit performance, we trains a model to map target performance metrics as inputs into circuit parameters as outputs.
%We hypothesize that, for a large variety of useful circuit topologies and a wide range of performance metrics, this inverse function is smooth and regular enough to learn from a relatively small number of examples obtained by circuit simulation.
%Supervised learning offers significantly higher data efficiency than reinforcement learning by focusing the training data on the relevant region of circuit parameter space, and does not require multiple simulator interactions during execution.

We address two inverse problems.
For the simpler one, described above and formalized in \cref{sec:prob1}, a dataset is created by simulating a circuit with parameter values on a grid that covers a user-specified range.
%The performance metrics measured by the simulator are stored in the dataset alongside the parameter values that induced them.
A neural network is then trained on this data to predict circuit parameters that would induce a desired performance vector.
We evaluate this approach on a much larger variety of useful circuit topologies than has previously been done, and show that this inverse function is smooth and regular enough to be approximated from a much smaller number of examples than achieved before, namely around 600–4000 points, depending on circuit complexity, compared with 10,000 to 40,000 points in prior work~\cite{datasize_1,datasize_2,datasize_3,datasize_4}.

However, this approach has severely limited usability, because it requires the user to make a rather precise guess of a feasible combination of performance metrics for the model to recover.
As the number of metrics of interest grows, in more complex circuits, the task of precisely specifying all metrics becomes daunting.
We instead envision an interface for a user to specify a vector of performance \emph{thresholds}, and propose a second inverse problem of mapping these thresholds into circuit parameters that satisfy them (\cref{sec:prob2}).
While this problem is natural for reinforcement learning algorithms \cite{2020autockt}, we propose a novel \emph{supervised learning} method for constructing, from the same simulation data as in the simpler problem, a dataset for training and evaluating a model that predicts threshold-satisfying parameters.
We show that training a neural network on this dataset solves this harder inverse problem an order of magnitude more efficiently than existing reinforcement learning methods, the latter using between 5500 and 40,000 simulations~\cite{2018learning,2020autockt}.
% \gc{In the next section, we will present the problem statement and the proposed approach, in Section 3, we will study the related work and compare their performance to our approach. In Section 4, the two deployed methods are explained and the experimental results are provided in Section 5. The paper is concluded in Section 6.}

This work contributes: (1) a novel and vastly more data-efficient method for generating, from circuit simulation data, a dataset for supervised learning of circuit design agents for the threshold specification problem; and (2) the to-date most extensive evaluation of automated circuit design methods on a diverse set of analog and radio frequency circuits, demonstrating the success of the method while also identifying a challenging circuit topology for future research. A demo of our proposed system is available at \href{circuits.streamlit.app}{circuits.streamlit.app}. 
%Our study demonstrates that the proposed method for automated circuit design is highly efficient and requires minimal data points for model training. 
%We compare previous methods with our work on a more diverse set of  analog and radio frequency circuits, which verifies that our method is versatile and applicable to various circuits.
% \rf{summarize contributions}

\section{Problem Statement}\label{sec:prob}

    Human design through the use of advanced electronic design automation (EDA) tools \cite{survey2} is currently the primary method for designing electronic circuits.
    However, human-led design is a slow process and is falling behind the human–computer co-design processes for digital circuits~\cite{genetic}.
In order to bridge the gap and allow for faster design of analog circuits, we aim to facilitate a system that can automatically generate the parameters of an analog circuit to meet a set of performance requirements.
A good system should be able to function with good accuracy across a variety of different circuit topologies.
In this paper, we therefore examine the problem of designing a diverse group of analog circuits, including single-stage amplifiers, multi-stage operational amplifiers, power amplifiers, low-noise amplifiers, nonlinear circuits such as mixers, and autonomous circuits such as voltage-controlled oscillators.
It is noteworthy that the selected performance metrics, themselves diverse across the various circuits, exhibit different kinds of correlations and tradeoffs. 

%\begin{figure}[!hbt]
%    \centering
%          \includegraphics[width=1.0\columnwidth]{fig/method.pdf}%
%          \label{fig:method}%
%    \caption{\rc{Combining supervised learning and reinforcement learning to benefit from both the high data efficiency of supervised learning that can reduce the required number of circuit simulations by up to three orders of magnitude; and from the autonomous exploration property of reinforcement learning that can efficiently search over large non-convex circuit design spaces.}}
%    \label{p2_nmos1}
%\end{figure}

%in the selected circuits to rule out any architecture-dependent performance.

% \begin{figure}[hpt]
%     \centering
%     \includegraphics[width=\columnwidth]{ProblemStatement.png}
%     \caption{Success percentage at different threshold\label{fig:method}}
%     \label{PS}
% \end{figure}

\subsection{Exact Specification}\label{sec:prob1}

For a specified circuit topology, let $n$ be the number of component parameters, such as resistances, transistor widths, and voltages.
Let $X_1, \ldots, X_n$ be the operational ranges of each of these parameters, and $X = \bigtimes_{i=1}^n X_i$ the design space.
We assume the availability of a simulator $f: X \to Y$, where $Y = \mathbb{R}_+^k$ is the positive orthant of the real vector space of $k$ performance metrics of interest.

The problem of design from exact specification is that of finding a function $g \approx f^{-1}: Y \to X$ such that, when a user specifies target performance $y \in Y$, the system can suggest a design $\hat{x} = g(y)$.
Upon suggesting $\hat{x}$, it can be simulated to measure its performance $\hat{y} = f(\hat{x})$.
The \emph{error} of the system is measured by the relative difference in its performance metrics
\begin{equation}\label{eq:acc1}
\delta_i = \frac{|y_i - \hat{y}_i|}{y_i}.
\end{equation}
For evaluation, the relative error is averaged across multiple test points as well as across the $k$ metrics.
% Since real-world applications can typically tolerate accuracy in the ballpark of 5\% before this source of error begins to dominate other sources \rf{cite}, we also report the system's \emph{success rate} as the fraction of test points with error within the $\epsilon \le 0.05$ margin, element-wise.
We also measure the \emph{success rate} as the fraction of test points with relative error within a given margin.

We note that, in a real-world system, users can input a target performance vector $y$ for which no circuit exists with low error.
The system can use the simulator to check that the predicted circuit $g(y)$ is incorrect, but it is a hard problem to determine whether another circuit would be correct, particularly if the instance is out-of-distribution for the data used to train the system.
We therefore focus on evaluating the system on in-distribution data $y \in f(X)$, and leave the challenging and interesting question of out-of-distribution generalization to future work.

\subsection{Threshold Specification}\label{sec:prob2}

%\bc{While being able to meet an exact set of performance metrics is a crucial step for automatic analog circuit design systems, there exists a larger problem we also wish to address. Many circuit design problems are a question of minimizing or maximizing the different performance metrics. As such a system must be able to address these requests.  Presently, some machine learning systems exist that can predict circuit parameters that create the exact performance metrics requested with reasonable accuracy \cite{mina2022,2015process,2016modelling,2017op,2018application,2018exploration,2020artificial,2021automated,2019power,2018learning,2020autockt} Further some prior works address the issue of thresholding certain metrics \cite{2018exploration}. However, we propose a novel method that achieves good accuracy while using less data than prior works. Shown in Figure \ref{PS} is the design-flow of our approach ...} 

When manual circuit design is challenging, guessing a feasible performance vector $y \in f(X)$ can be just as challenging, particularly if it consists of many metrics that are subject to intricate tradeoffs.
Instead, it would be easier for a user to specify performance thresholds that the designed circuit should meet.
We denote by $\lambda_i$ the threshold direction of metric $i$, i.e. $\lambda_i = 1$ or $-1$ respectively whether it is majorative (the more the better) or minorative (the less the better).

\begin{figure}[t]
    \centering
          \includegraphics[width=1.0\columnwidth]{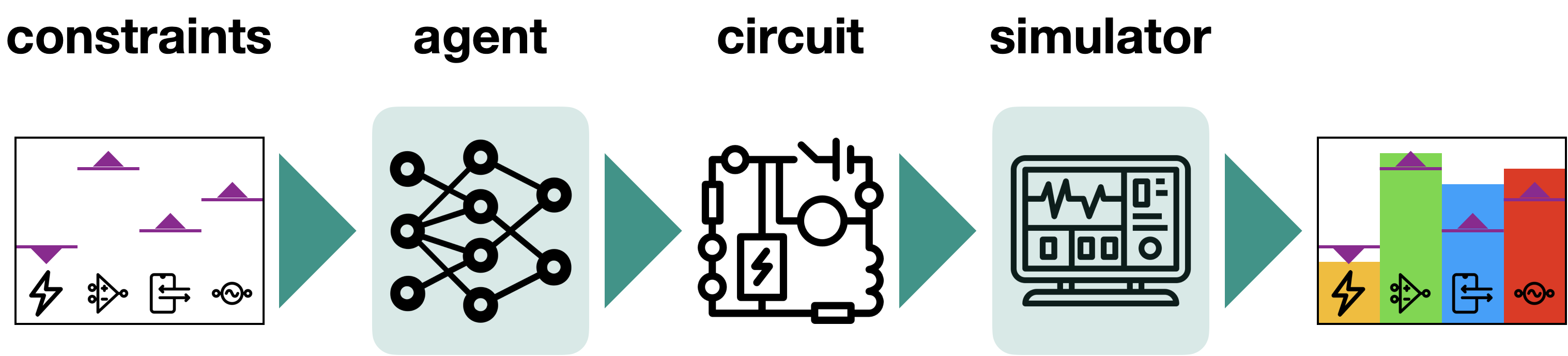}%
    \caption{The problem of automated design by threshold specification. A user specifies threshold constraints on the circuit's performance metrics. A design agent then generates a circuit that, when simulated, meets the constraints.\label{fig:problem}}
\end{figure}

The problem of design from threshold specification (\cref{fig:problem}) is that of finding a function $g: Y \to X$ such that, when a user specifies target performance thresholds $y \in Y$, the suggested design $\hat{x} = g(y)$ aims to meet the thresholds $y$ by having its simulated performance $\hat{y} = f(\hat{x})$ satisfy $\lambda \hat{y} \ge \lambda y$ element-wise.
The error of this system is measured by the relative amount of threshold violation
\begin{equation}\label{eq:acc2}
\delta_i = \frac{\max\{\lambda_i(y_i - \hat{y}_i), 0\}}{y_i}.
\end{equation}
As before, we measure success rate by the fraction of test data for which the  thresholds for all metrics are met up to a given error margin.

To evaluate a system solving the threshold specification problem, we should use threshold queries that follow a similar distribution to that of real users.
Leaving user studies to future work, we approximate this distribution by perturbing simulated performance metrics similarly to \citet{2018exploration}.
Given the measured performance $y = f(x)$ of a simulated circuit $x$, we sample standard uniform perturbations $u \sim \mathrm{U}^k$ for the $k$ metrics, independent and identically distributed (i.i.d), and use the perturbed vector
\begin{equation}\label{eq:eps}
\tilde{y}_i = (1 - \epsilon \lambda_i u_i) y_i
\end{equation}
as the threshold query.
Here $\epsilon$ is the perturbation magnitude hyperparameter; in this work we use $\epsilon = 0.2$.
Note that, by construction, $\lambda y \ge \lambda \tilde{y}$, so that there always exists a circuit (namely, $x$) that meets the threshold $\tilde{y}$.

\section{Related Work}

\subsection{Digital Circuits vs. Analog Circuits}
Digital circuit automation and computer-assisted design~(CAD) has progressed steadily over the past few decades \cite{digitalcad1, digitalcad2}.
The invariant architecture of the building blocks in digital design allows the application of graph-theoretic approaches that treat the problem of digital circuit design as a graph connectivity problem, which has led to a large body of work in optimization of digital design \cite{optim1, optim2,optim3}.
Analog circuits, on the other hand, involve a set of unique design challenges that are not considered in the digital domain.
First, analog circuits have a broad range of architectures, and each building block may be optimized individually with respect to a performance metric before all the blocks are integrated into the circuit.
Second, in digital design, there is a small set of critical performance metrics and in most cases only power consumption, area, and speed are considered.
In contrast, in the analog domain, a variety of performance metrics are present, and optimization of an analog circuit becomes a higher-dimensional problem.
Third, in analog circuits, passive components such as capacitors, inductors, and resistors are also deployed, completely changing the dynamics of the circuit design and weakening the relation between graph-theoretic properties and circuit performance. %This is the major reason that in this paper, we focus on the design of analog and RF circuits. 

\subsection{Automated Analog Circuit Design}

Automating the design of analog circuits has been studied before, particularly in operational amplifiers (op-amps) that are specified by their voltage gain, bandwidth, and power consumption (for a survey, see \citet{mina2022}).
% It is noteworthy \rf{why? does this somehow explain why these are the most popular metrics?}, that one of the most well-known trade-offs in analog circuits is the gain-bandwidth product of an amplifier which comes from the device-centric frequency limitations, i.e., the $f_T$ \cite{ft}.\gc{This inherent trade-off between performance metrics of an amplifier impacts the sample space by ...}
\citet{2018learning} propose a reinforcement learning (RL) approach to designing 3-stage amplifier circuits from threshold specification.
Similarly, \citet{2020autockt} adopt RL to design 2-stage operational amplifiers.
While RL is readily amenable to threshold constraints, it suffers from poor data efficiency compared with supervised learning approaches~\cite{mina2022}.
\citet{2015process} use supervised regression to design another type of circuits, a 4-bit current-steering Digital-to-Analog converters (DAC), from exact specification of the performance metrics.
Other works have used supervised learning to design various op-amps~\cite{2020artificial,2018exploration,2021automated} with varying — and often incomparable — data efficiency~\cite{mina2022}.

In this paper, we step beyond the scope of op-amp design to additionally investigate the design of other critical analog circuit blocks, in particular radio-frequency electronic circuits that are commonly used in cellular communication applications \cite{razavi}. It is noteworthy that some of the selected circuits, e.g., mixers and oscillators, are among the most nonlinear analog circuits with high sensitivity to variations in design parameters.
We further show that design agents for amplifiers as well as more intricate circuits can be learned by supervised regression from much smaller datasets than previously accomplished.
Finally, we learn to design these circuits from threshold specification, in contrast to most previous supervised learning works.
\citet{2018exploration} previously considered this setting, and proposed a method that we reproduce in this paper under the name $D^m_\epsilon $.
We show that this method can lead to suboptimal performance, analyze the reason through an ablation study, and propose a new method that mitigates this issue.

\section{Method}

\begin{figure}[t]
    \centering
          \includegraphics[width=1.0\columnwidth]{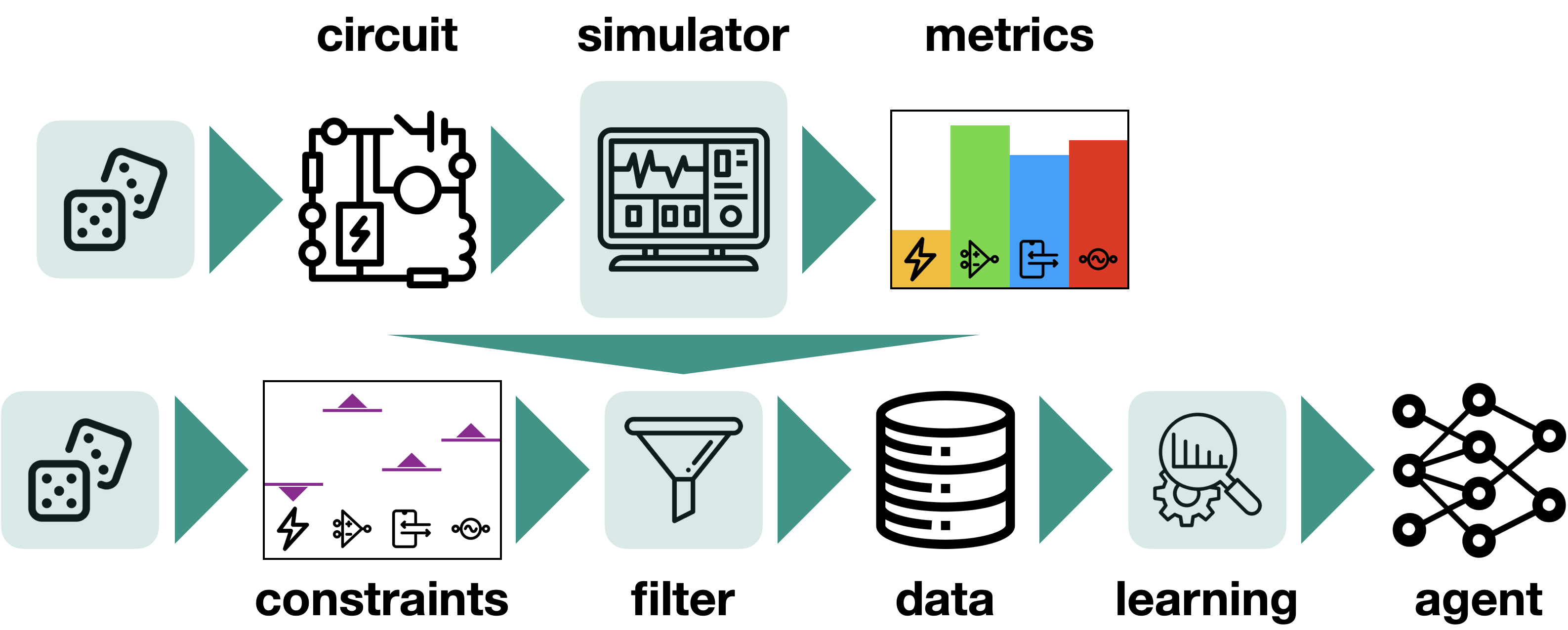}%
    \caption{Proposed method for automated design from threshold specification. Circuit parameters are sampled within a user-defined range, simulated, and measured. Performance metrics are randomly adjusted to sample threshold queries. A data filtering process then generates training data for supervised learning a circuit design agent that generates circuits to meet threshold requirements.}
    \label{fig:method}
\end{figure}

We use supervised learning to approximate the inverse of the simulator function mapping circuit parameters to performance metrics (\cref{fig:method}).
We interface an external simulator to generate
%NgSpice takes in a netlist and some parameters in order to generate a dataset used for training. 
a dataset $D_0$ consisting of circuit parameter vectors $x \in \mathbb{R}^n$ and their respective measured performance metrics vectors $y \in \mathbb{R}^k$.
We (optionally) pass this dataset through a filtering pipeline that prepares it for solving the threshold specification problem (\cref{sec:prob2}).
Finally, we employ a supervised learning algorithm, such as gradient-based optimization, to train a design agent.
%We then use the Adam optimization algorithm~\cite{adam} to train a neural network that takes in the performance metrics $y$ and outputs a prediction of what circuit parameters $\hat{x}$ would achieve the target performance $y$.
In this section we describe the system components: the simulator, the agent model, and several alternatives for the filtering pipeline. 
%The next section details the evaluation experiments and their results.
%Once trained, the network to take in any performance for the circuit it was trained for and output the parameters that make the requested performance. 

\subsection{Simulator}\label{sec:sim}

In this work, we use the NgSpice simulator \cite{ngspice}.
The circuit topology and its fixed parameters, as well as the simulation parameters, are provided to the simulator via a format called netlist \cite{netlist}.
In addition to the netlist, the simulator loads analysis commands that determine how it measures the performance metrics of interest.
For some circuits, multiple analysis commands are given to measure the circuit under distinct conditions.

The external simulator is wrapped by a Python interface to allow easy access to two functionalities.
First, to generate simulation data, a user inputs the range and step size of each circuit parameter, and the simulator loops through this grid to output a dataset $D_0$ of parameter–metrics pairs.
Second, to evaluate the trained model, predicted circuit parameters are input to the simulator, and the measured performance is compared with the target performance.
%, and it can be con-
%trolled by command window which means having well interface with python. Netlists and parameter
%ranges are determined by virtuoso. Reformat the netlist in a .sp file and use ngpsice language to edit
%analysis section. There are two files which are training and testing files for each circuit. Training
%file Generates the data for network to learn, and the parameter ranges will be provided. Testing file
%check the accuracy of parameter predictions.Ngspice can load the circuit file which contains the netlist and analysis command, and it can be controlled by command window which means having well interface with python.
%
%\aqc {In addition to loading circuit files that contain netlists and analysis commands, NGSPICE can be controlled by command windows, which implies a well-integrated Python interface. Netlists and parameter ranges are determined by Cadence Virtuoso.  An analysis section has been added to the netlist by reformatting it for NGspice. Testing analog circuits may require several test benches; similarly, different performance metrics  analyses may require different netlists. Upon simulation results approval, the netlists are applied for training.}

\subsection{Agent Model}\label{sec:model}

Before the raw data from the simulator can be put through the model, we apply a few data pre-processing steps.
The different features of the data have vastly different scales.
In order to allow the model to learn across such different scales, we first shift and scale all values to the range $[-1,1]$.
%In more detail, we scale each feature in the training data (both parameters and performance measures) according to their highest and lowest values. This way all the features have consistent range and values, enabling faster learning. 
This normalization is applied both to the performance metrics before they are fed to the model and to the ground-truth circuit parameters used for training, and an appropriate inverse operator is applied to the model's parameter predictions.  

In this work, we experiment with three different agent models.
The main model is a neural network with an architecture of a simple multi-layer perceptron, trained with the Adam optimizer~\cite{adam}.
%The experiments described in Section 5 involve differing numbers of circuit parameters and performance metrics.
The network takes in a vector of desired performance metrics and predicts a vector of circuit parameters, which is then compared with the ground-truth parameters using an absolute ($L_1$) loss.
The sizes of the first and last layers of the network are adjusted to reflect the number of performance metrics and circuit parameters, respectively, which are different for each experiment described in \cref{sec:exp}.
The architecture is otherwise constant across experiments and detailed in Appendix \ref{appendix_a}.
An alternative model we consider is ensembles of decision trees trained with the Random Forests algorithm~\cite{breiman2001random}.
Finally, to assess the need for any learning at all, we compare with a lookup method that memorizes the training data and selects, for each test performance vector, the training circuit that minimizes the relative performance error.

\subsection{Filtering Pipeline}

The problem of design from exact specification (\cref{sec:prob1}) can be solved by supervised learning, in which the training set is the simulation dataset $D_0$, inverted so that performance metrics $y$ are inputs and circuit parameters $x$ are outputs.
However, this method is unlikely to be sufficient for the threshold specification problem (\cref{sec:prob2}), in which some threshold vectors are out-of-distribution for $D_0$, because no circuit has them as its exact performance.
We therefore propose a filtering pipeline that constructs, from the same $D_0$, a second dataset which, when used for supervised learning, trains a model that predicts circuit parameters from threshold specification.

To prepare a circuit for the threshold specification problem, two properties of the metrics vector need to be provided.
First, because some metrics, such as gain or bandwidth, are majorative (the more the better), while others, such as power consumption, are minorative (the less the better), we need to know for each metric $i$ its threshold direction $\lambda_i \in \{-1, 1\}$.
Second, a specification asking for the highest gain at power consumption at most $p$ is different from one asking for the lowest power consumption that achieves gain at least $g$.
We may therefore have a preference order over metrics, such that we lexicographically prefer improving $y_i$ over improving $y_j$, whenever $i < j$, as long as all threshold constraints are approximately met.
We say that $y$ is lexicographically better than $y'$ if there exists $i$ such that $y_j = y'_j$ for all $j < i$ and $\lambda_i y_i > \lambda_i y'_i$.

The filtering pipeline starts by finding, for each performance vector $y \in D_0$, all \emph{feasible} performance vectors $y' \in D_0$ that meet the threshold specification $y$, i.e.
\begin{equation}
F(y; D_0) = \{(x, y') \in D_0 | \lambda y' \ge \lambda y\}.
\end{equation}
The design agent needs to map the threshold specification $y$ to one such $x \in F(y)$, but it may not be immediately clear which one.
We hypothesize that, crucially to learning with high success rate from small datasets, our training dataset must be systematic in selecting a representative of $F(y)$.
This systematicity manifests as a pattern that the learning algorithm can generalize, whereas including the entire $F(y)$ or selecting from it sporadically might lead to conflicts that impede generalization.

We propose to select the lexicographically best training-set circuit that meets the threshold
\begin{equation}
\bar{D}^*_0 = \{(x, y) | y \in D_0, x = \argmax F(y; D_0)\},
\end{equation}
where we select for $x$ a single representative $(x, y')$ of $F(y)$ that maximizes $y'$ lexicographically.
In the notation $\bar{D}^*_0$, the bar denotes feasibility of $x$ for $y$ and the star denotes selection of the best representative.
%With $(x_0, y'_0)$ this representative, we add $(x_0, y)$ to $D'$. {\color{yellow} Maybe we should clarify this method more. For example, why can't we just add $(x_0, y')$}

We note that, by definition, all members of $F(y)$ have good circuit parameters that meet the threshold $y$.
However, adding all of them to our training set, similar to the method proposed by~\citet{2018exploration}, would create conflicts where the same network input $y$ is mapped to different outputs. 
%{\color{teal} We examine the effects of having these conflicts in the method titled All Epsilon Feasible in the results.}
By breaking “ties” in a consistent way — and in accordance with user-specified preference over metrics — we create a dataset more conducive of learning.
The new dataset $\bar{D}^*_0$ has the same size as the simulation dataset $D_0$ and the same set of performance vectors. 
The circuit parameter vectors in $\bar{D}^*_0$ are those that define its Pareto frontier, that is, for which no other simulated circuit is better in all performance metrics.
Thus, $\bar{D}^*_0$ consistently maps feasible performance vectors to frontier circuits.
%{\color{teal} We refer to this dataset construction method as Best Feasible in later sections.}

%{\color{yellow} Here we need to add information about SoftArgmax, Ablation and Lourenco methods}

\subsubsection{Threshold Queries}

In \cref{sec:prob2}, we discussed how performance metrics measured in simulation are perturbed to generate threshold queries (Eq. (\ref{eq:eps})).
We denote thus perturbed data by
\begin{equation}
D_\epsilon = \{ (x, (1 - \epsilon \lambda u)y) | (x, y) \in D_0, u \sim U^k \text{ i.i.d} \}.
\end{equation}
Note that the distribution of threshold queries $y \sim D_\epsilon$ is different than the distribution of simulated metrics vectors $y \sim \bar{D}^*_0$.
To avoid a mismatch of the training and test distributions, we combine the filters to form a dataset of threshold queries with a principled selection of target circuits:
\begin{equation}\label{eq:method}
\bar{D}^*_\epsilon = \{ (x, \tilde{y}) | \tilde{y} \in D_\epsilon, x = \argmax F(\tilde{y}; D_0) \}.
\end{equation}
$\bar{D}^*_\epsilon$ is a dataset mapping $\epsilon$-perturbed metrics vectors $\tilde{y}$ to circuits whose (unperturbed) simulated metrics are feasible for the threshold query $\tilde{y}$, selecting the lexicographically best such circuit.

\subsubsection{Baseline and Ablation}

We compare our dataset construction methods, $\bar{D}^*_0$ and $\bar{D}^*_\epsilon$, with a baseline that closely follows~\citet{2018exploration}.
We define $D^m_\epsilon$ as the union of $m$ i.i.d. samples of $D_\epsilon$
\begin{equation}
D^m_\epsilon = \bigcup_{t=1}^m D_\epsilon[u_t]; \quad u_t \sim U^k \text{ i.i.d}.
\end{equation}
In our experiments, $m = 20$.
The reasons are that by construction, in each $(x, \tilde{y}) \in D^m_\epsilon$ the circuit $x$ is feasible for the threshold query $\tilde{y}$, i.e. $\lambda f(x) \ge \lambda \tilde{y}$ ;
and that the training distribution $\tilde{y} \sim D^m_\epsilon$ is identical to our evaluation distribution $\tilde{y} \sim D_\epsilon$.
Note that, in contrast to most of the literature on analog circuit design automation via supervised learning, which employs a simulation dataset akin to $D_0$, $D^m_\epsilon$ is suited for the threshold specification problem~\cite{2018exploration}.

Unfortunately, the dataset $D^m_\epsilon$ can be very confusing to learn from.
Because the simulator function $f$ is not necessarily injective, there may exist multiple circuits with similar performance vectors.
Moreover, such vectors have overlapping supports of their perturbation distributions.
The result is that $D^m_\epsilon$ will tend to have similar threshold queries mapped to vastly different circuit parameters, rendering their prediction difficult.

We propose an ablation that more directly demonstrates this issue.
In $\bar{D}^m_\epsilon$, we select for each $\tilde{y} \in D_\epsilon$ the $m$ lexicographically-best feasible circuits, rather than only the single best in $\bar{D}^*_\epsilon$ (Eq. (\ref{eq:method})):
\begin{equation}
\bar{D}^m_\epsilon = \{ (x, \tilde{y}) | \tilde{y} \in D_\epsilon, x \in \topm F(\tilde{y}; D_0) \}.
\end{equation}
We expect this method to perform suboptimally, more similarly to $D^m_\epsilon$ than to $\bar{D}^*_\epsilon$.
This would provide evidence that the main aspect impacting the prior method, compared with the novel one, is the existence of multiple targets for each query, rather than the other differences — namely, the selection of circuits from the feasible set $F(y)$, or the preference of lexicographically better circuits.

To summarize, we consider six datasets: (1) $D_0$ is the simulation data; (2) $D_\epsilon$ has perturbed performance metrics that resemble the threshold query distribution, and is used for method evaluation; (3) $\bar{D}^*_0$ and (4) $\bar{D}^*_\epsilon$ are our proposed methods, without and with perturbation to match the test distribution; (5) $D^m_\epsilon$ is a baseline similar to~\citet{2018exploration}; and (6) $\bar{D}^m_\epsilon$ is an ablation study.

\section{Experiments}\label{sec:exp}

We experiment with our methods on a diverse group of seven circuit topologies, detailed below. Best practices in circuit design suggest that circuit parameters are chosen based on their impact on performance metrics ~\cite{Bandler1988,hassan2016novel,Bandler2023}. Only these parameters are used to optimize performance for each circuit.
We simulate each circuit in a parameter grid consisting of approximately 4000 points, except for the simplest two-stage amplifier with around 600 points, as presented in Table \ref{train-data} in Appendix \ref{appendix_f}.
A schematic of each circuit shows the range and step size of each variable parameter, as well as color-coded tags illustrating the diversity of the circuits to which our method applies. To facilitate result reproduction, the code and data used in our experiments are available at github. \footnote{https://github.com/indylab/Circuit-Synthesis} The supplementary details of the circuits employed in our experiments can be found in Tables \ref{range_1}, \ref{range_2}, and \ref{range_3} in Appendix \ref{appendix_f}.

Our main method uses the $\bar{D}^*_\epsilon$ dataset to train a neural network and evaluate its success rate in 10-fold cross-validation.
For each circuit topology, we perform three comparisons of this method.
First, we compare the main method with the five other data construction methods described in the previous section.
Second, we compare the gradient-based learning algorithm with Random Forests and a simple lookup method~(\cref{sec:model}).
Third, %after an exploratory search for the amount of simulation data that is sufficient to successfully train a network for each circuit, 
we study the sensitivity to the amount of training data by varying it.
We compare the success rate of 10-fold cross validation, which uses 90\% of the data for training each fold, with using 5\%, 10\%, 20\% and 50\% of the data for training.
We do this by randomly splitting the data into (respectively) 20, 10, 5, and 2 disjoint subsets, training on one subset, testing on the rest, and then averaging the result across the splits.

In all plots in this section, the solid curve is the average over 10  runs of data splitting and training, and the shaded area is the standard-error of the mean (SEM) over those runs.

\subsection{Analog Voltage Amplifiers}

\subsubsection{Common Source (CS) Amplifier}

\begin{figure}[t]
 %  \centering
   \includegraphics[width=1\columnwidth]{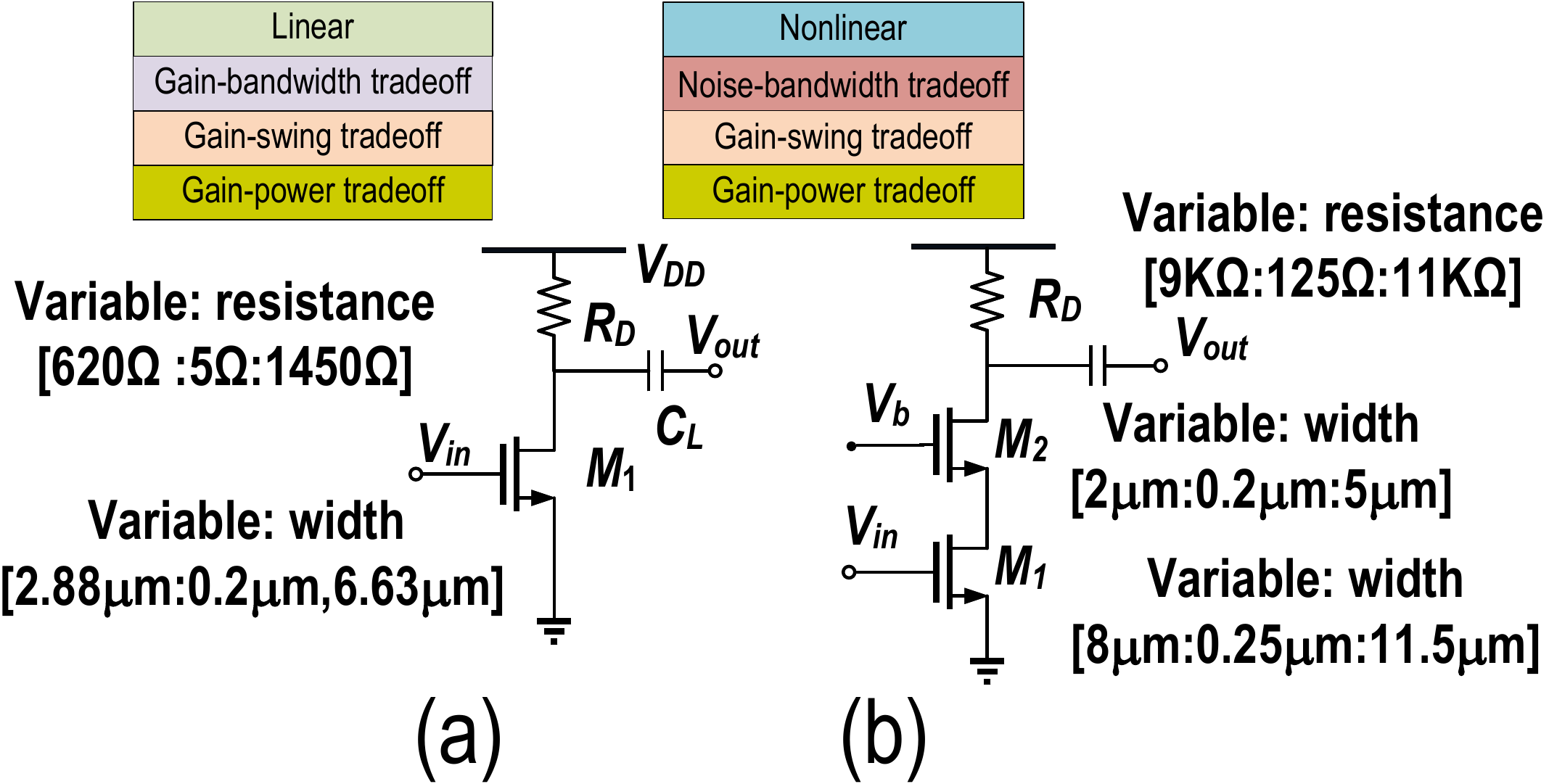}
   \caption{Schematics of analog voltage amplifiers: (a) common source (CS) amplifier; (b) cascode amplifier.  Color-coded tags show circuit characteristics. Ranges and step sizes are marked near each circuit parameter.}
    \label{nmos_cascode}
\end{figure}

Due to its simplicity, the common source (CS) amplifier~(\cref{nmos_cascode}(a)) is among the most popular amplifier configurations using a CMOS transistor.
As design variables, we consider the width of the transistor and the resistance of the load resistor $R_D$.
The target performance metrics, in decreasing importance are: bandwidth, voltage gain, and power consumption.

% \begin{figure*}[!htb]
%    \begin{minipage}{1\columnwidth}
%      \centering
%      \includegraphics[width=0.33\linewidth]{fig/p2_datasize/nmos/test-margin_SoftArgmax_new.png}
%      \caption{Interpolation for Data 1}\label{Fig:Data1}
%    \end{minipage}\hfill
%    \begin{minipage}{1\columnwidth}
%      \centering
%      \includegraphics[width=0.33\linewidth]{fig/p2/nmos/subset-new.png}
%      \caption{Interpolation for Data 2}\label{Fig:Data2}
%    \end{minipage}
%     \begin{minipage}{1\columnwidth}
%      \centering
%      \includegraphics[width=0.33\linewidth]{fig/p2/nmos/subset-new.png}
%      \caption{Interpolation for Data 2}\label{Fig:Data2}
%    \end{minipage}
% \end{figure*}

% \begin{figure*}[!hbt]
% \begin{minipage}[t]{0.3\textwidth}
%   \includegraphics[width=\linewidth]{fig/p2_datasize/nmos/test-margin_SoftArgmax_new.png}
%   \caption{Common Source Amplifier, comparing datasizes for BestEpsilonFeasible}
%   \label{fig:first}
% \end{minipage}%
% \hfill % maximize the horizontal separation
% \begin{minipage}[t]{0.3\textwidth}
%   \includegraphics[width=\linewidth]{fig/p2/nmos/subset-0.9-margin.png}
%   \caption{Common Source Amplifier, comparing Datasets}
%   \label{fig:second}
% \end{minipage}%
% \hfill
% \begin{minipage}[t]{0.3\textwidth}
%   \includegraphics[width=\linewidth]{fig/p2_methods/nmos/subset-0.9-margin.png}
%   \caption{Common Source Amplifier, comparing Methods}
%   \label{fig:third}
% \end{minipage}%
% \end{figure*}

\begin{figure}[t]
\centering
\hfill
\subfigure[]{\label{res:cs1}\includegraphics[width=0.48\columnwidth]{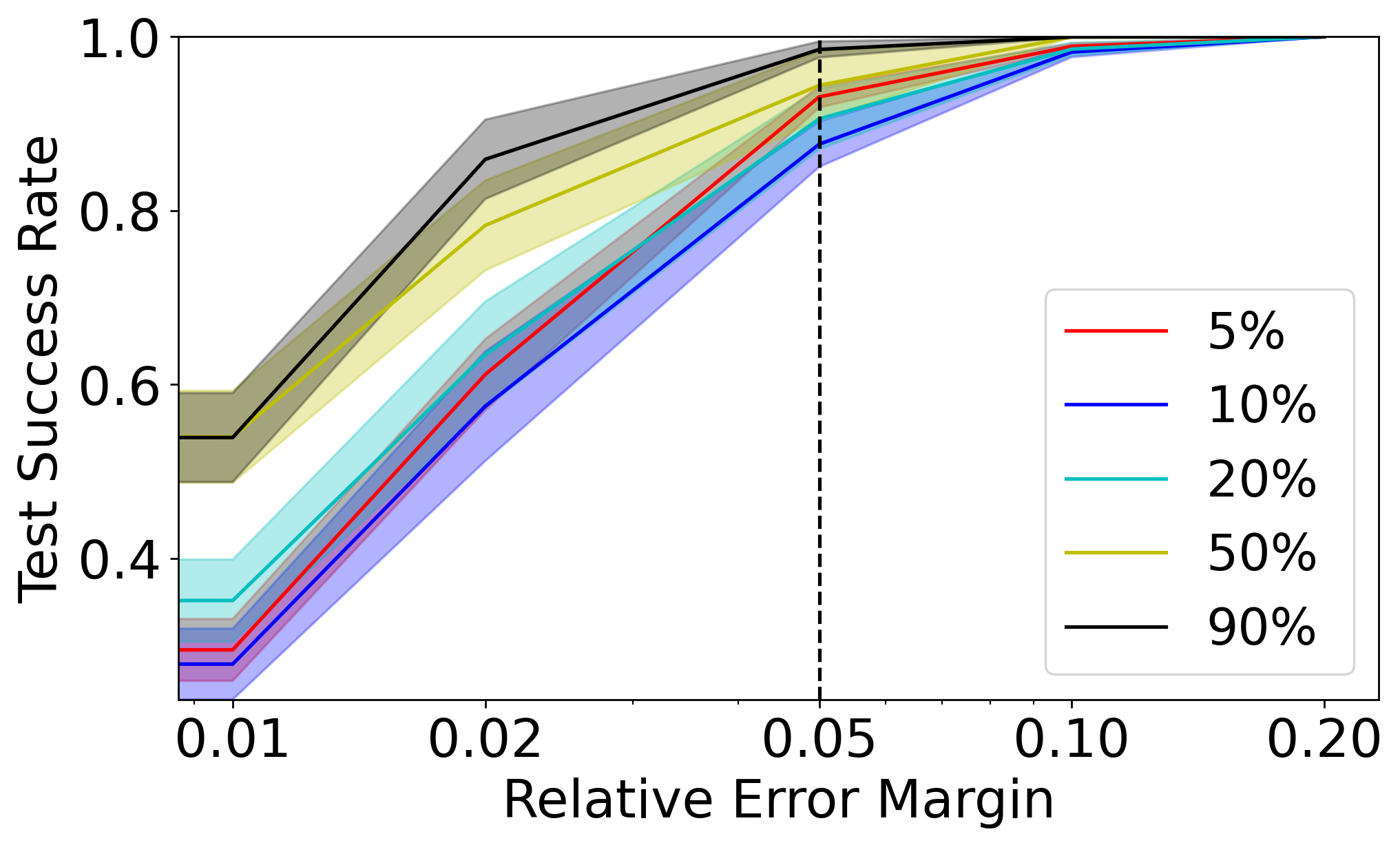}} \hfill
\subfigure[]{\label{res:cs2}\includegraphics[width=0.48\columnwidth]{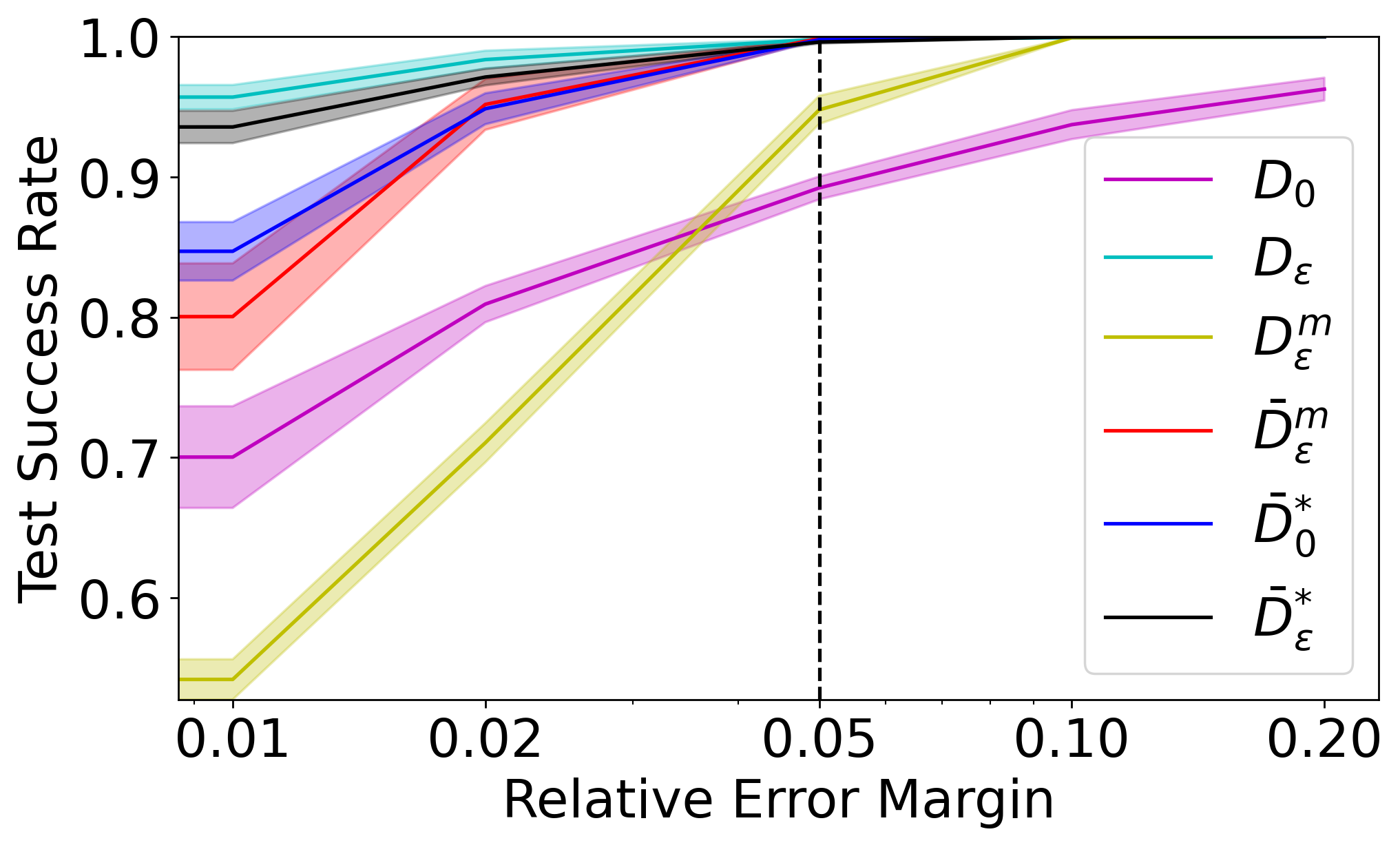}}
\hfill
\caption{Success rate of designing a CS amplifier: \textbf{(a) Exact specification:} training on $D_0$ with varying data sizes and testing on exact metrics $y \sim D_0$; \textbf{(b) Threshold specification:} training on varying datasets and testing on threshold metrics $y \sim D_\epsilon$.\label{res:cs}}
\end{figure}

As shown in \cref{res:cs1}, our model achieves near-perfect success at 5\% error margin on the exact specification problem~(\cref{sec:prob1}), even while using 6 times less data than the best previous work \cite{Devi2021}.
In the threshold specification problem (\cref{sec:prob2}), our model trained on the $\bar{D}^*_\epsilon$ dataset also achieves perfect success at 5\% error margin, whereas training on the naïve $D_0$ baseline dataset only achieves 85\% success.
Note, however, that all other data processing methods also achieve perfect success on this simple circuit.
Further results appear in Appendix \ref{appendix_c}.
%\yc{ In the following Figure \ref{p2_nmos1}, using only 5 percent of the training data, the model can achieve accuracy close to 90 percent with a margin of 5 percent}. The two best methods, BestFeasible and BestEpsilonFeasible perform similarly. Random Forest and Neural Network outperformed 1 - Nearest Neighbour by a gap of 5 percent using 90 percent of the data.  

\subsubsection{Cascode Amplifier}

The CS amplifier has limited gain and exhibits a trade-off between critical performance metrics.
The cascode amplifier shown in \cref{nmos_cascode}(b) enhances the amplification bandwidth compared with a CS stage \cite{cascode}.

\begin{figure}[t]
\centering
\hfill
\subfigure[]{\label{res:cascode1}\includegraphics[width=0.49\columnwidth]{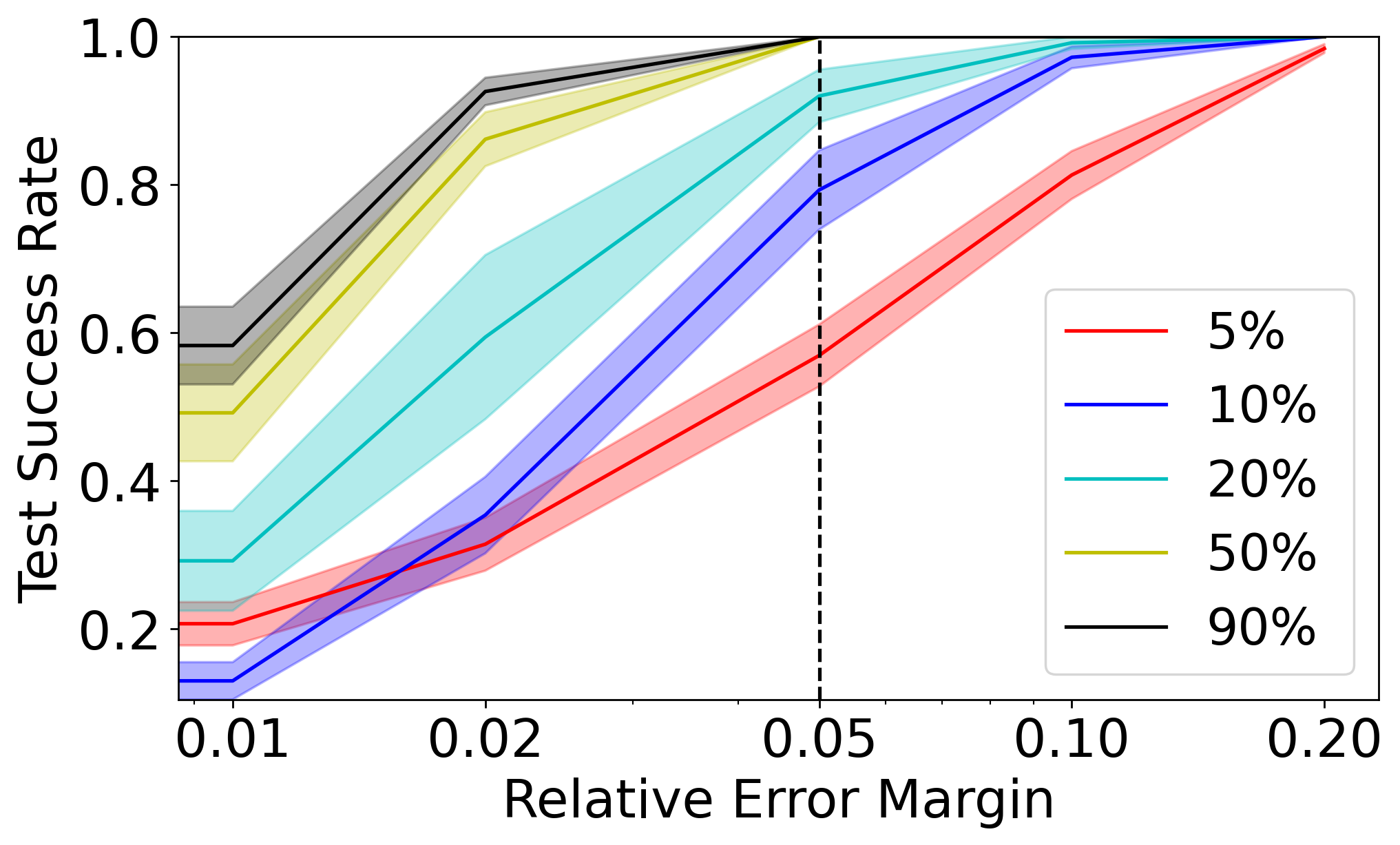}} \hfill
\subfigure[]{\label{res:cascode2}\includegraphics[width=0.49\columnwidth]{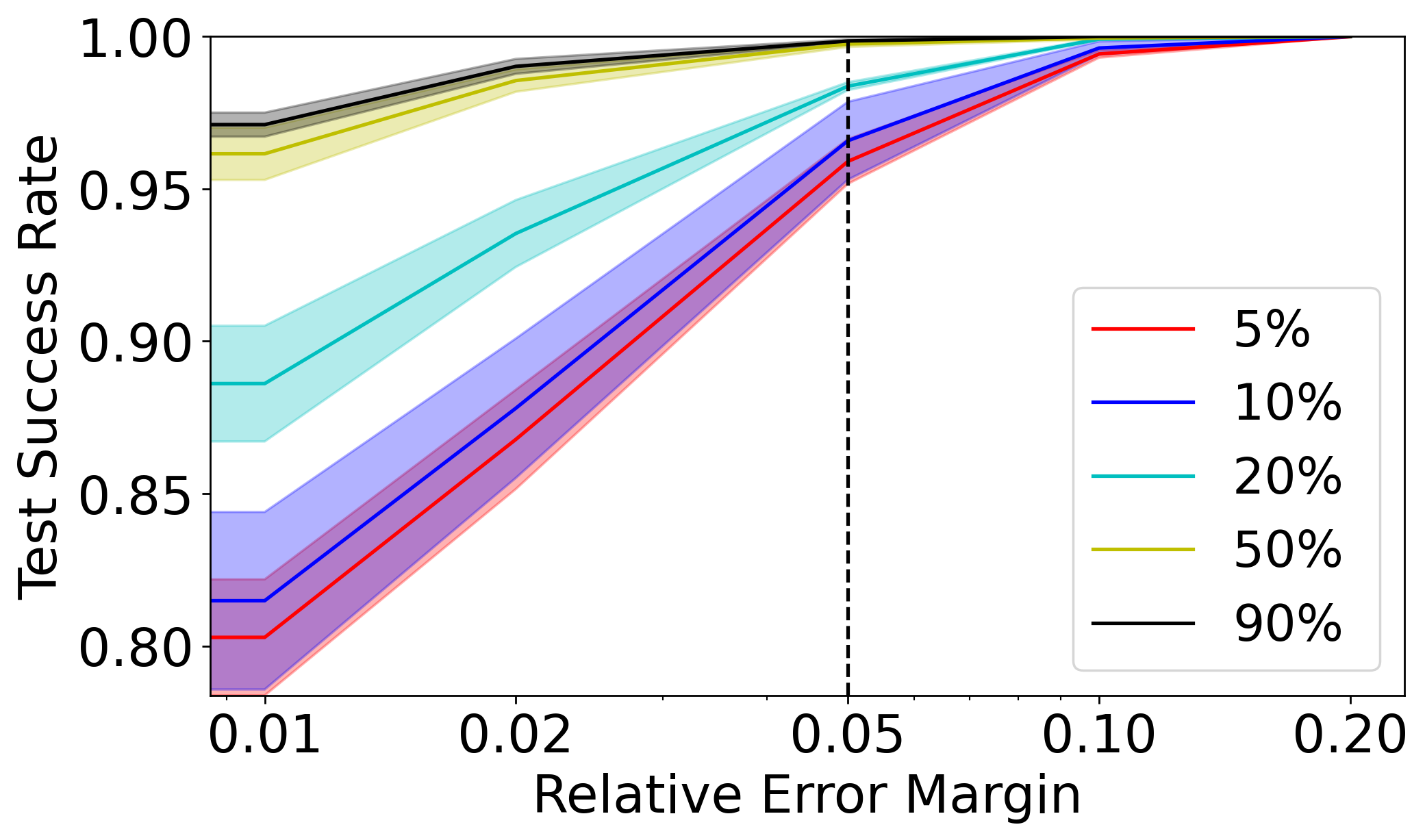}} \hfill
\hfill \\ \hfill
\subfigure[]{\label{res:cascode3}\includegraphics[width=0.48\columnwidth]{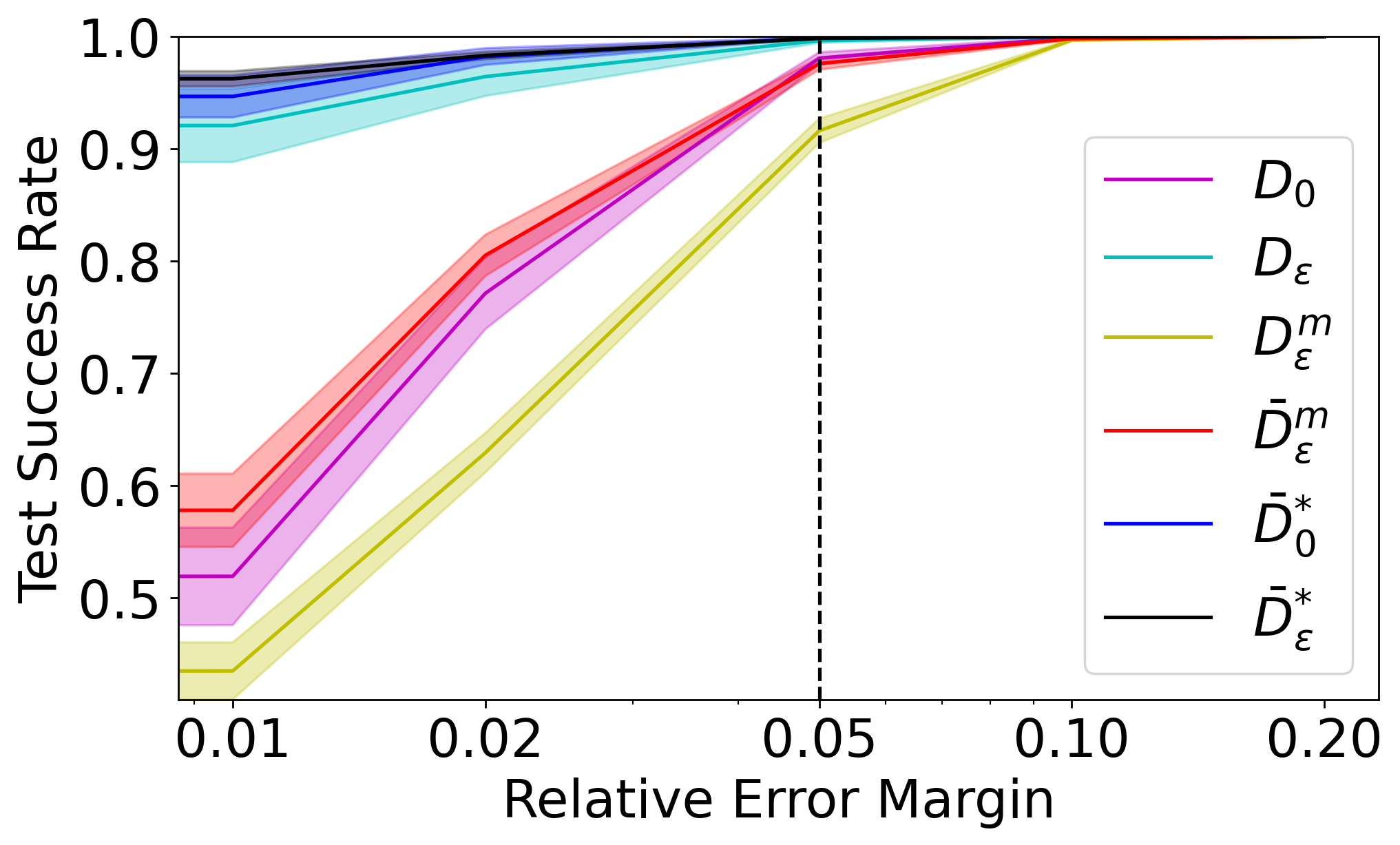}} \hfill
\subfigure[]{\label{res:cascode4}\includegraphics[width=0.48\columnwidth]{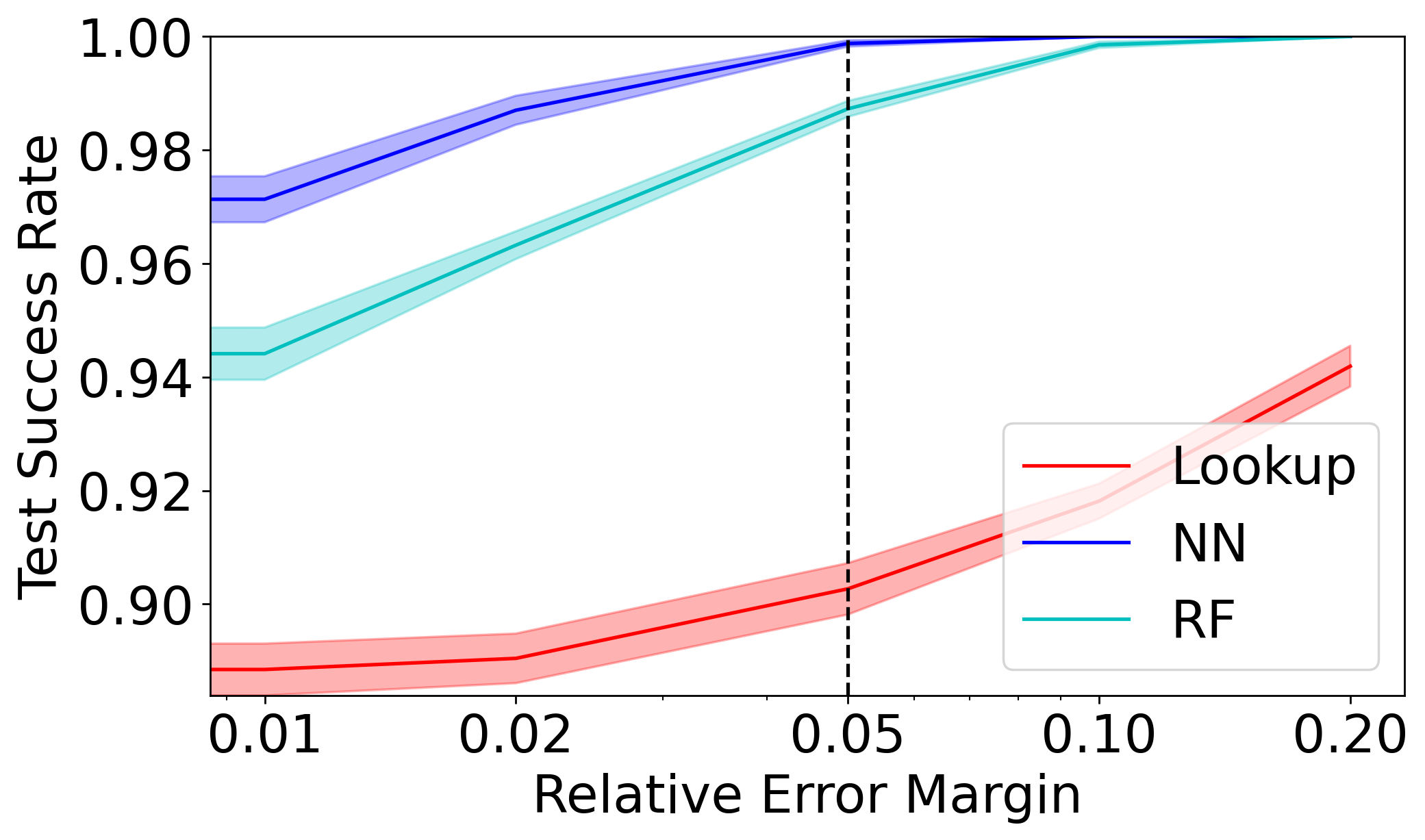}} \hfill
\hfill
\caption{Success rate of designing a cascode amplifier: \textbf{(a) Exact specification}; \textbf{(b–d) Threshold specification}, comparing: (b) different data sizes of the $\bar{D}^*_\epsilon$ dataset; (c) different datasets; and (d) different agent models.\label{res:cascode}}
\end{figure}

%\begin{figure}[!hbt]
%    \centering
%          \includegraphics[width=.8\columnwidth]{fig/p2/cascode/subset-0.9-margin.png}%
%          \label{fig:left}%
%    \caption{Cascode, comparing datasets}
%    \label{p2_cascode}
%%    \vspace{-0.15in}
%\end{figure}

% \yc{For the Cascode Amplifier, our model learns the mapping from 3 performance variables to 3 parameters}. 

As illustrated in \cref{res:cascode}, this more challenging circuit shows more sensitivity to the amount of training data, both in (a) the exact specification and (b) the threshold specification settings.
It is also more difficult for the baseline methods to achieve high success rate, particularly when a low error margin is needed: the simulation dataset $D_0$ and the “non-injective” datasets $D^m_\epsilon$ and $\bar{D}^m_\epsilon$ all tend to generate circuits with higher than 1\%, 2\%, and rarely even 5\% threshold violation (\cref{res:cascode3}).
Finally, \cref{res:cascode4} demonstrates the typical underperformance of the lookup method, showing that learning is needed; while also demonstrating an uncommon case where Random Forests slightly underperforms the neural network.
%Unlike the CS amplifier, the baseline method shows a more significant gap of almost 50 percent. We can also see that EpsilonBase performs comparably to our method. 

% \begin{table}[!hbt]
%  \small
% % \singlespacing
%  \centering
% \caption{Cascode Circuit Parameter Training Info}\label{table1}
% \begin{tabular}{|p{3cm}|p{1cm}|p{1cm}|p{1cm}|}
% \hline
% \textbf{Parameter/Info} & \textbf{r} & \textbf{w0} & \textbf{w1}\\
% \hline
% Parameter start value & 200 & 4u & 7u \\
% \hline
% Parameter end value & 500 & 7.5u & 10u \\
% \hline
% Parameter step value & 18.75 & 0.25u & 0.2u \\
% \hline

% \end{tabular}
% \end{table}

% \begin{figure}[!hbt]
%     \centering
%        \subfloat[Average Error]{%
%           \includegraphics[width=5cm]{Cascode_Margin_resize.png}%
%           \label{fig:left}%
%        } 
%        \subfloat[5 percent validation margin accuracy per epochs]{%
%           \includegraphics[width=5cm]{Cascode_accuracy_resize.png}%
%           \label{fig:middle}%
%        }
%        \subfloat[Validation Loss per epochs]{%
%           \includegraphics[width=5cm]{cascode_loss_resize.png}%
%           \label{fig:right}%
%        }
%        \caption{Overview performance of Cascode circuit}
%        \label{fig:default}
% \end{figure}

\subsubsection{Two-Stage Amplifier}
\begin{figure}[t]
 \centering
\includegraphics[width=0.92\columnwidth]{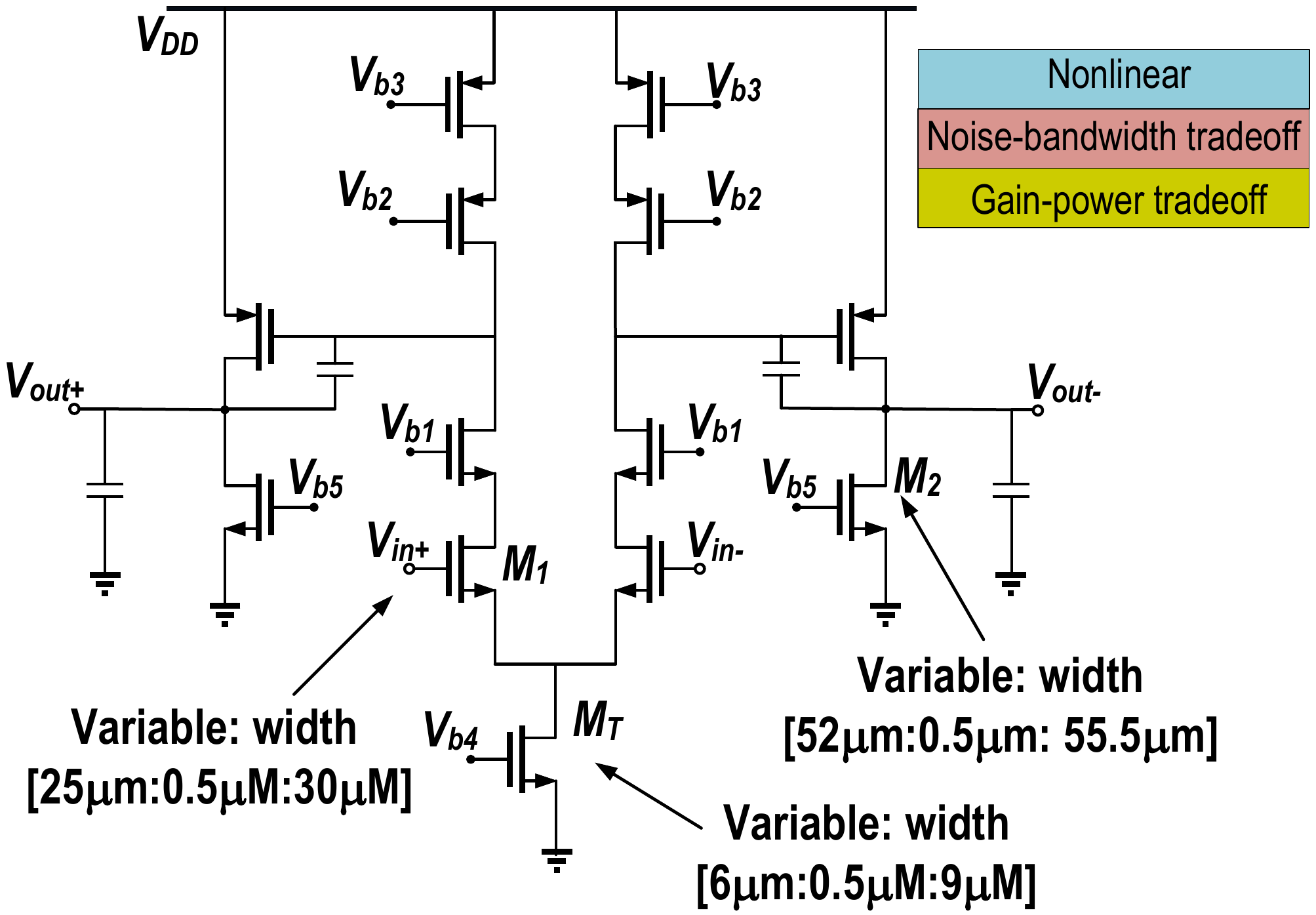}
    \caption{Schematic of a two-stage amplifier.}
    \label{twostage}
     \end{figure}

The circuits of \cref{nmos_cascode} both suffer from limited gain.
Two-stage amplifiers, as shown in \cref{twostage}, are excellent replacements of single-stage amplifiers and in particular allow simultaneously achieving higher gain and voltage swing~\cite{meyer}.

\begin{figure*}[t]
\centering
\hfill
\subfigure[]{\label{res:2s}\includegraphics[width=0.19\textwidth]{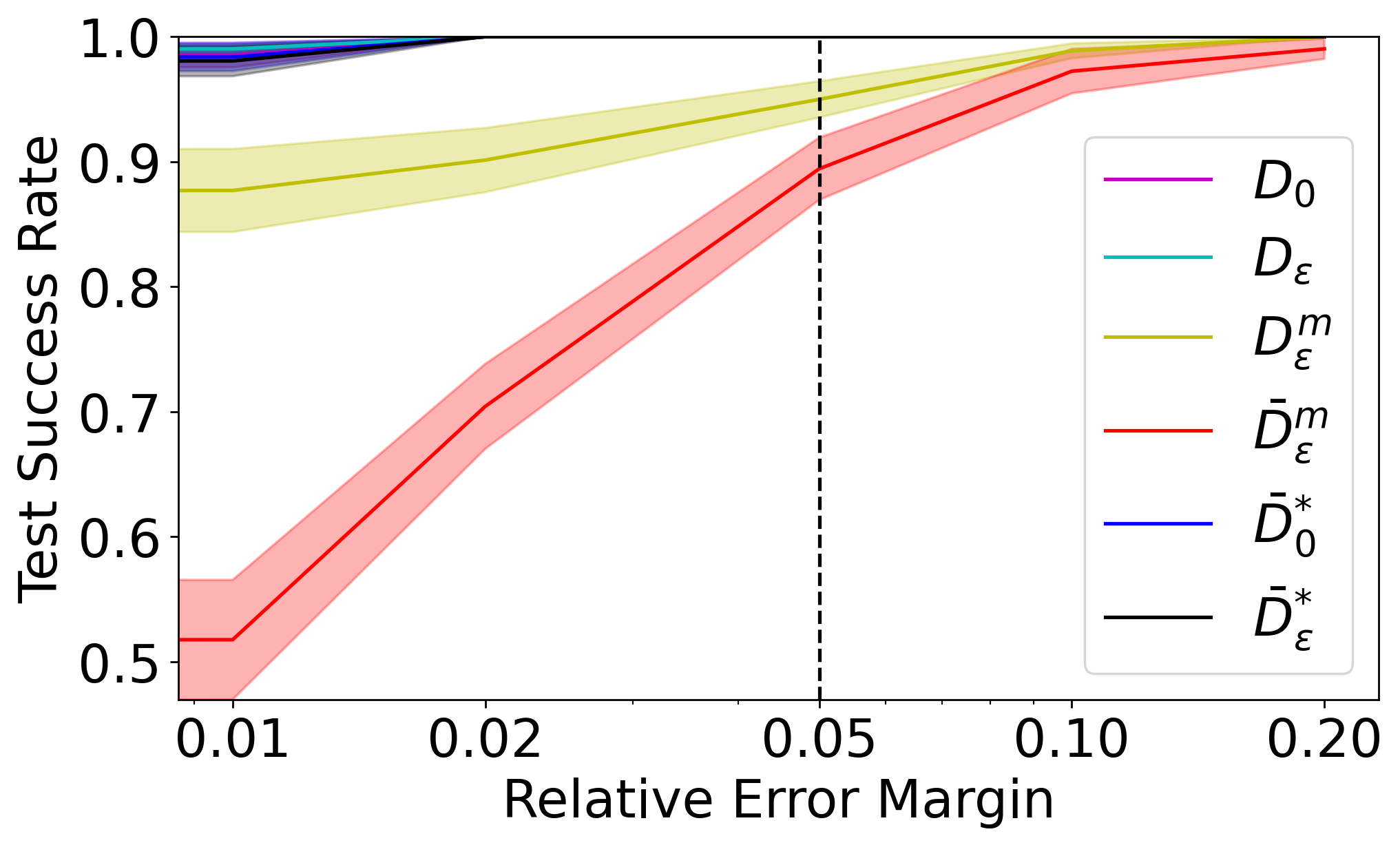}} \hfill
\subfigure[]{\label{res:lna}\includegraphics[width=0.195\textwidth]{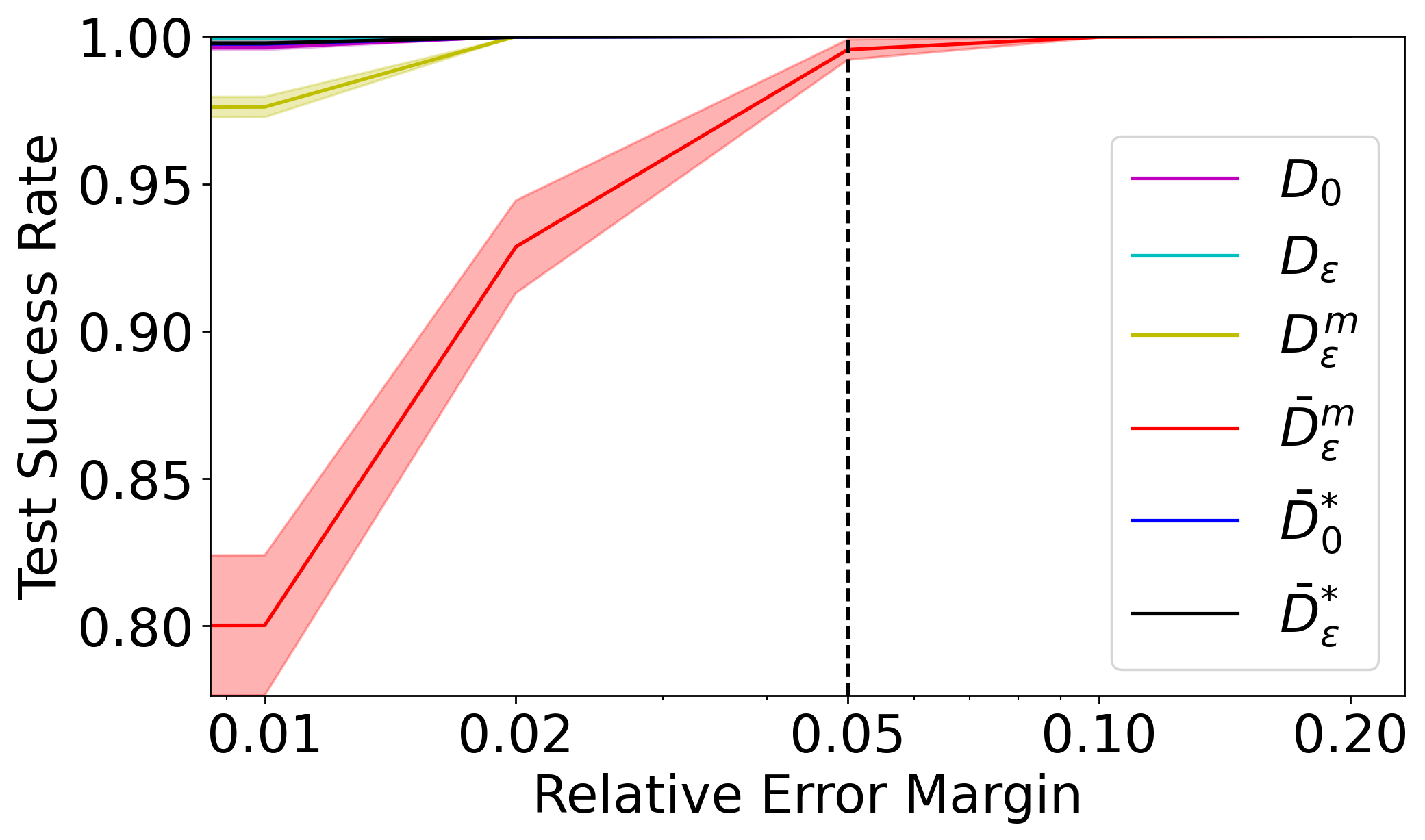}} \hfill
\subfigure[]{\label{res:pa}\includegraphics[width=0.19\textwidth]{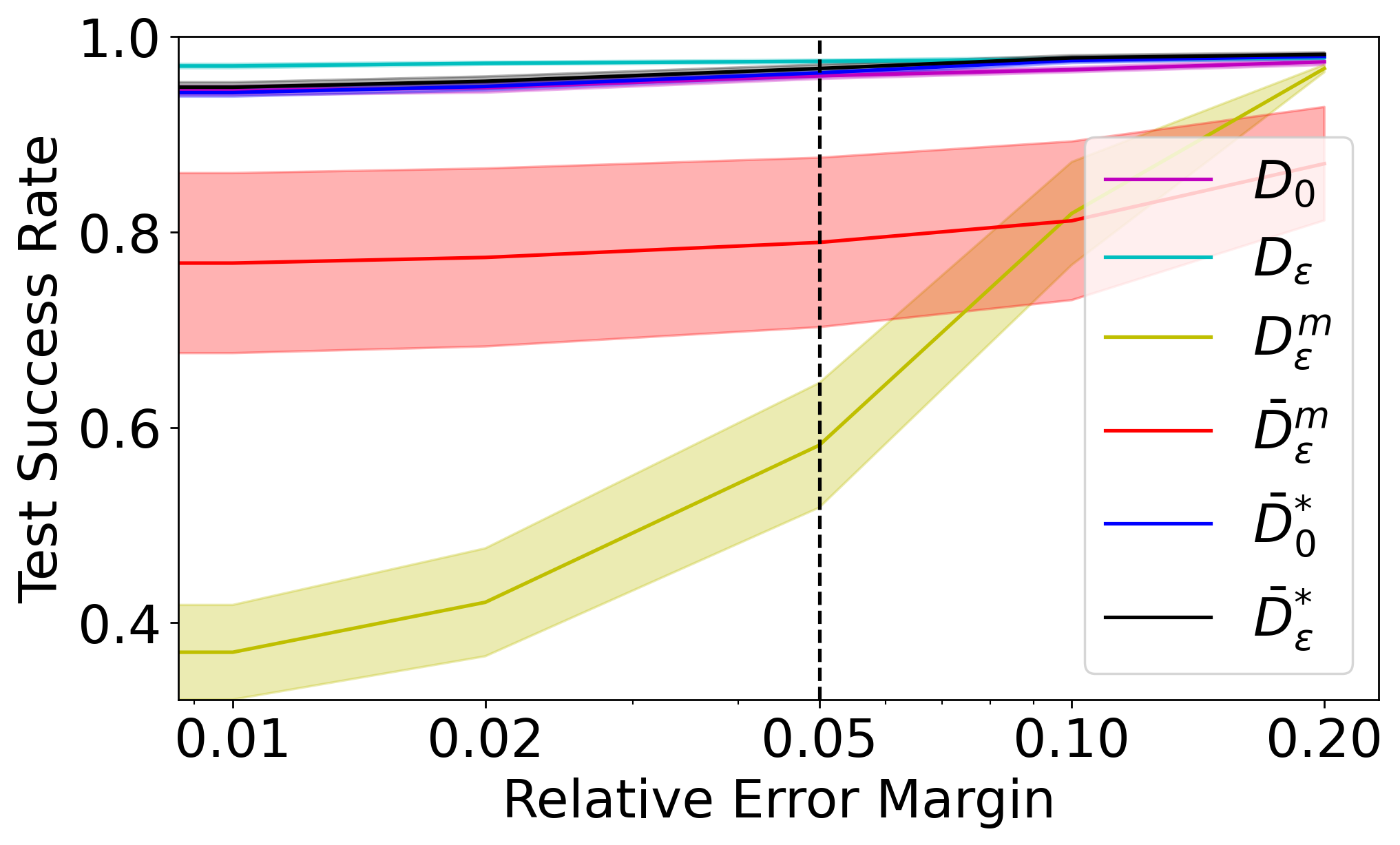}} \hfill
\subfigure[]{\label{res:mixer}\includegraphics[width=0.19\textwidth]{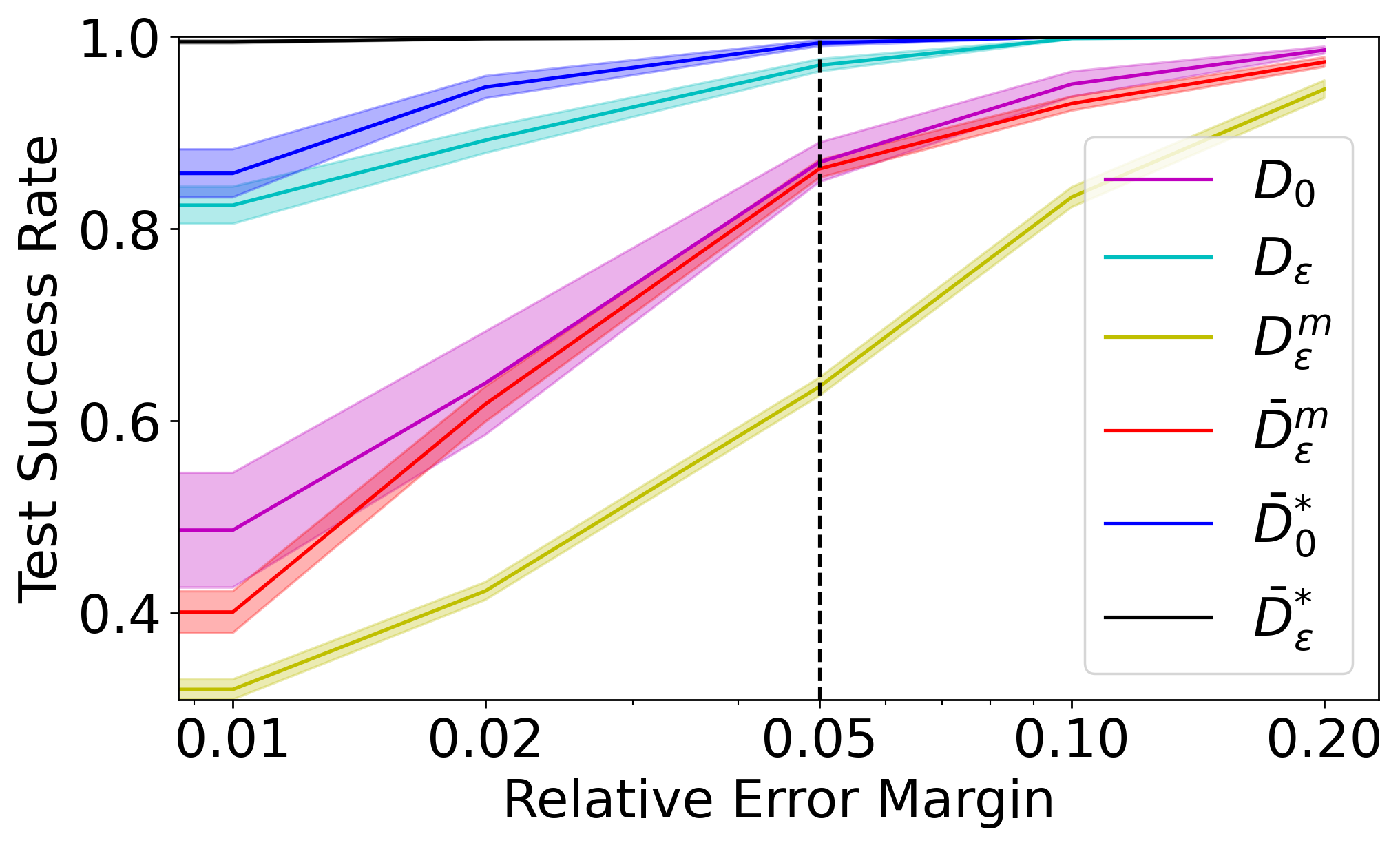}} \hfill
\subfigure[]{\label{res:vco}\includegraphics[width=0.19\textwidth]{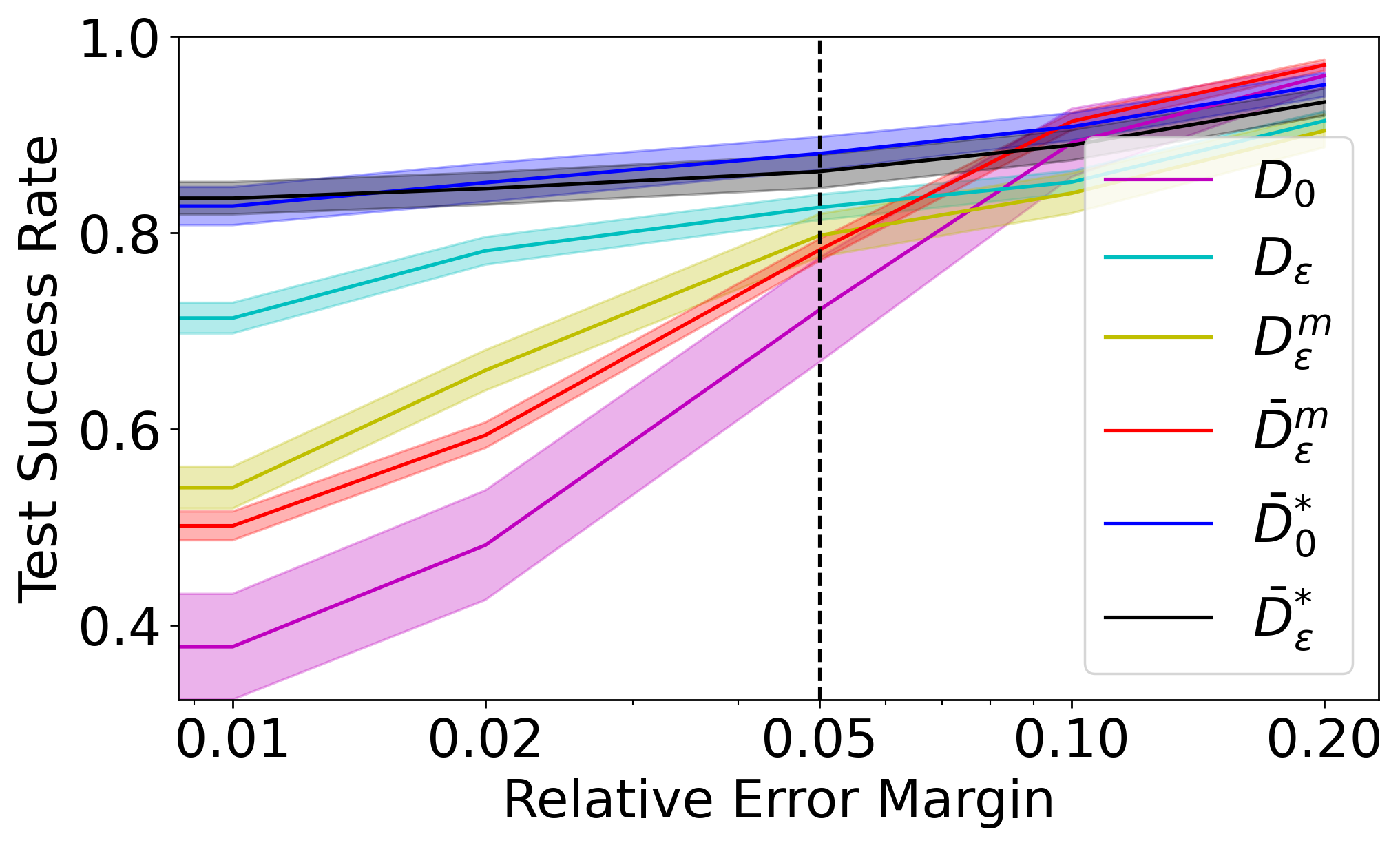}} \hfill

\caption{Comparing success rate in the threshold specification problem for different training datasets in five circuits: (a) two-stage; (b) low-noise amplifier (LNA); (c) power amplifier (PA); (d) mixer; (e) voltage-controlled oscillator (VCO).
\label{res:others}}
\end{figure*}

%\begin{figure}[t]
%    \centering
%          \includegraphics[width=.8\columnwidth]{fig/p2/two_stage/subset-0.9-margin.png}%
%          \label{fig:left}%
%    \caption{Two Stage , comparing datasets}
%    \label{p2_two_stage1}
%%     \vspace{-0.15in}
%\end{figure}

% \begin{table}[!hbt]
%  \small
% % \singlespacing
%  \centering
% \caption{Two Stage Circuit Parameter Training Info}\label{table1}
% \begin{tabular}{|p{3cm}|p{1cm}|p{1cm}|p{1cm}|}
% \hline
% \textbf{Parameter/Info} & \textbf{w0} & \textbf{w1} & \textbf{w2}\\
% \hline
% Parameter start value & 25u & 6u & 52u \\
% \hline
% Parameter end value & 30u & 9u & 55.5u \\
% \hline
% Parameter step value & 0.5u & 0.5u & 0.5u \\
% \hline
% \ha{the parameter values are listed in the figures, do we still need these tables?} 

% \end{tabular}
% \end{table}

% \begin{figure}[!hbt]
%     \centering
%        \subfloat[Average Error]{%
%           \includegraphics[width=5cm]{Two_Stage_Margin_Resize.png}%
%           \label{fig:left}%
%        } 
%        \subfloat[5 percent validation margin accuracy per epochs]{%
%           \includegraphics[width=5cm]{Two_stage_Accuracy_Resize.png}%
%           \label{fig:middle}%
%        }
%        \subfloat[Validation Loss per epochs]{%
%           \includegraphics[width=5cm]{Two_stage_loss_resize.png}%
%           \label{fig:right}%
%        }
%        \caption{Overview performance of two stage circuit}
%        \label{fig:default}
% \end{figure}

Owing to their widespread use, two-stage amplifiers have been among the most popular benchmark circuit configurations examined in prior automation work~\cite{mina2022,2020autockt}.
%; therefore we evaluate our method on this circuit as well.}
% \yc{Two stage amplifier like the Cascode amplifier requires learning  $f(R^3) \rightarrow  R^3$.}
Other than the baseline dataset $D^m_\epsilon$ and ablation dataset $\bar{D}^m_\epsilon$, all other datasets achieve perfect success on this relatively easy circuit (\cref{res:2s}), using 10 times less data than in the best previous work \cite{datasize_1,datasize_2,datasize_3,datasize_4}.
The underperformance of the “non-injective” datasets supports our hypothesis that a systematic selection of representative circuits for similar performance levels is needed to facilitate learning of circuit design agents for threshold specification.
%Stage amplifier achieves similar performance to the BestEpsilonFeasible method. KEpsilonFeasible and AllEpsilonFeasible  have a large gap of 25 and 45 percent compared to BestEpsilonFeasible. }

\subsection{(Non)-Linear Radio Frequency Circuits}
\subsubsection{Low-Noise Amplifier (LNA)}

\begin{figure}[t]
    \centering
    \includegraphics[width=.95\columnwidth]{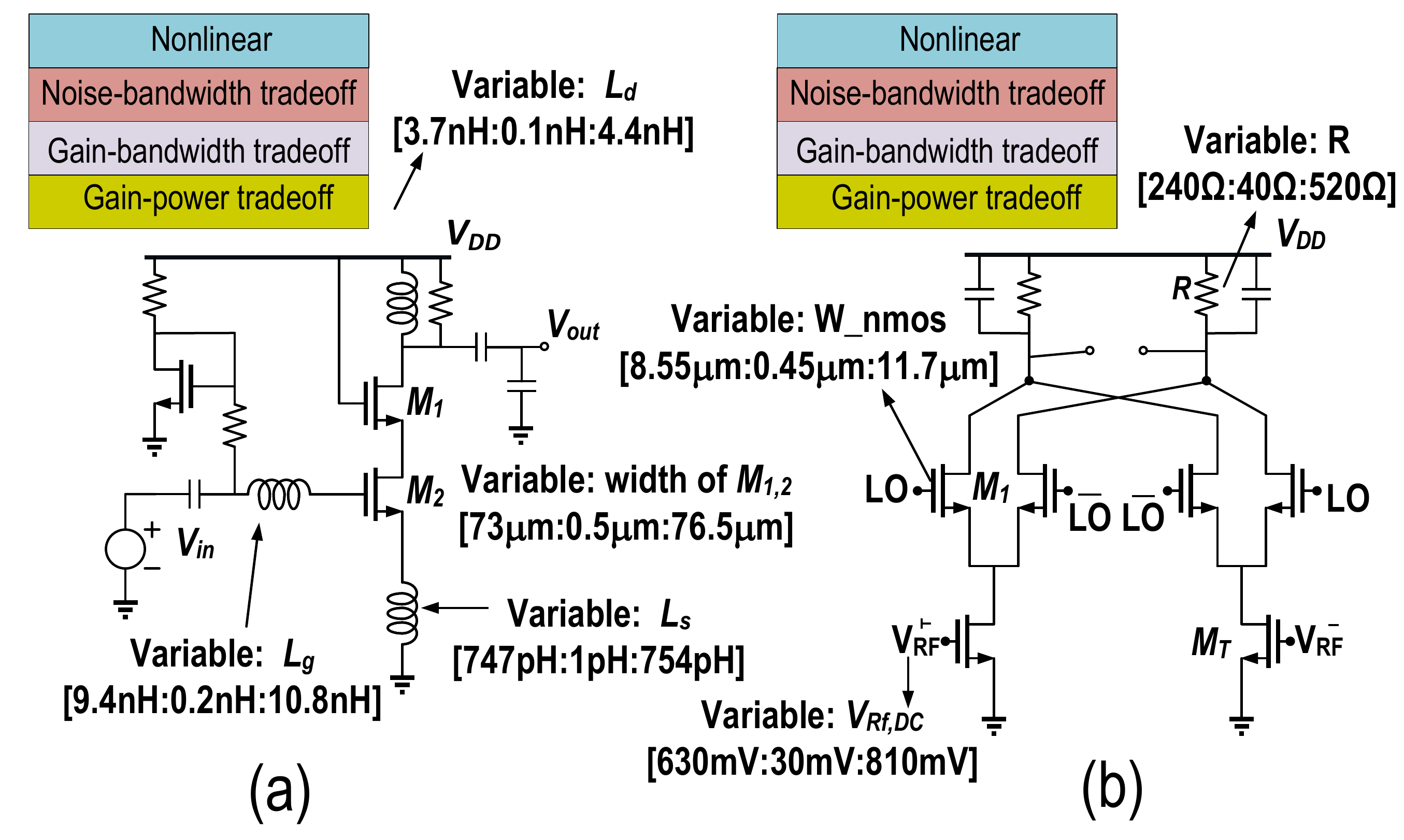}
    \caption{Schematics of: (a) low noise amplifier (LNA); (b) mixer.}
    \label{circuit_LNA_Mixer}
\end{figure}

The cascode low-noise amplifier (LNA) with inductive degeneration is a popular configuration to design an LNA for an RF receiver~\cite{LNASD}. The circuit, depicted in \cref{circuit_LNA_Mixer}(a), can obtain a high gain and minimal loss of input power across a large bandwidth without suffering from additive noise of circuit components (mainly transistors).
This circuit has four parameters (three inductor values and one cascode transistor width) and three metrics: the noise figure (signal-to-noise ratio between the input and output), the return loss, and the power gain.

% \yc{$f(R^3) \rightarrow  R^4$ }

Our findings indicate that the LNA simulation function has a smooth surface, making it easy to invert for all methods, including the baseline, and resulting in perfect success even at very low error margins (\cref{res:lna}).
%Since the LNA circuit performance requirement can be easily met \rc{why is it easy?}, all the methods  including the Baseline achieve almost 100\% test success rate at 0.5\% - 1\% margin. 
Only the ablation method $\bar{D}^m_\epsilon$ performs suboptimally, suggesting that it is even more prone to inconsistencies than the baseline $D^m_\epsilon$.
%We hypothesize that during the data modification phase, \rc{some data with aliasing is generated} which makes the task hard to learn.}

%\begin{figure}[!hbt]
%    \centering
%          \includegraphics[width=0.8\columnwidth]{fig/p2/lna/subset-0.9-margin.png}%
%          \label{fig:left}%
%    \caption{LNA, comparing datasets}
%    \label{LNA1}
%%    \vspace{-0.15in}
%\end{figure}

% \begin{table}[!hbt]
%  \small
% % \singlespacing
%  \centering
% \caption{LNA Circuit Parameter Training Info}\label{table1}
% \begin{tabular}{|p{3cm}|p{1cm}| p{1cm}| p{1cm}| p{1cm}|}
% \hline
% \textbf{Parameter/Info} & \textbf{ls} & \textbf{ld}  & \textbf{lg}  & \textbf{w}\\
% \hline
% Parameter start value & 55p & 3.8n & 14n & 49u \\
% \hline
% Parameter end value & 62p & 6.2n & 17n & 54u 

% \hline
% Parameter step value & 0.6p & 0.6n & 0.4n & 0.5u \\
% \hline

% \end{tabular}
% \end{table}

% \begin{figure}[!hbt]
%     \centering
%        \subfloat[Average Error]{%
%           \includegraphics[width=5cm]{LNA_margin_resize.png}%
%           \label{fig:left}%
%        } 
%        \subfloat[5 percent validation margin accuracy per epochs]{%
%           \includegraphics[width=5cm]{LNA_accuracy_resize.png}%
%           \label{fig:middle}%
%        }
%        \subfloat[Validation Loss per epochs]{%
%           \includegraphics[width=5cm]{LNA_loss_resize.png}%
%           \label{fig:right}%
%        }
%        \caption{Overview performance of LNA Circuit}
%        \label{fig:default}
% \end{figure}

\subsubsection{Power Amplifier (PA)}
\begin{figure}[hpt]
    \centering
    \includegraphics[width=.95\columnwidth]{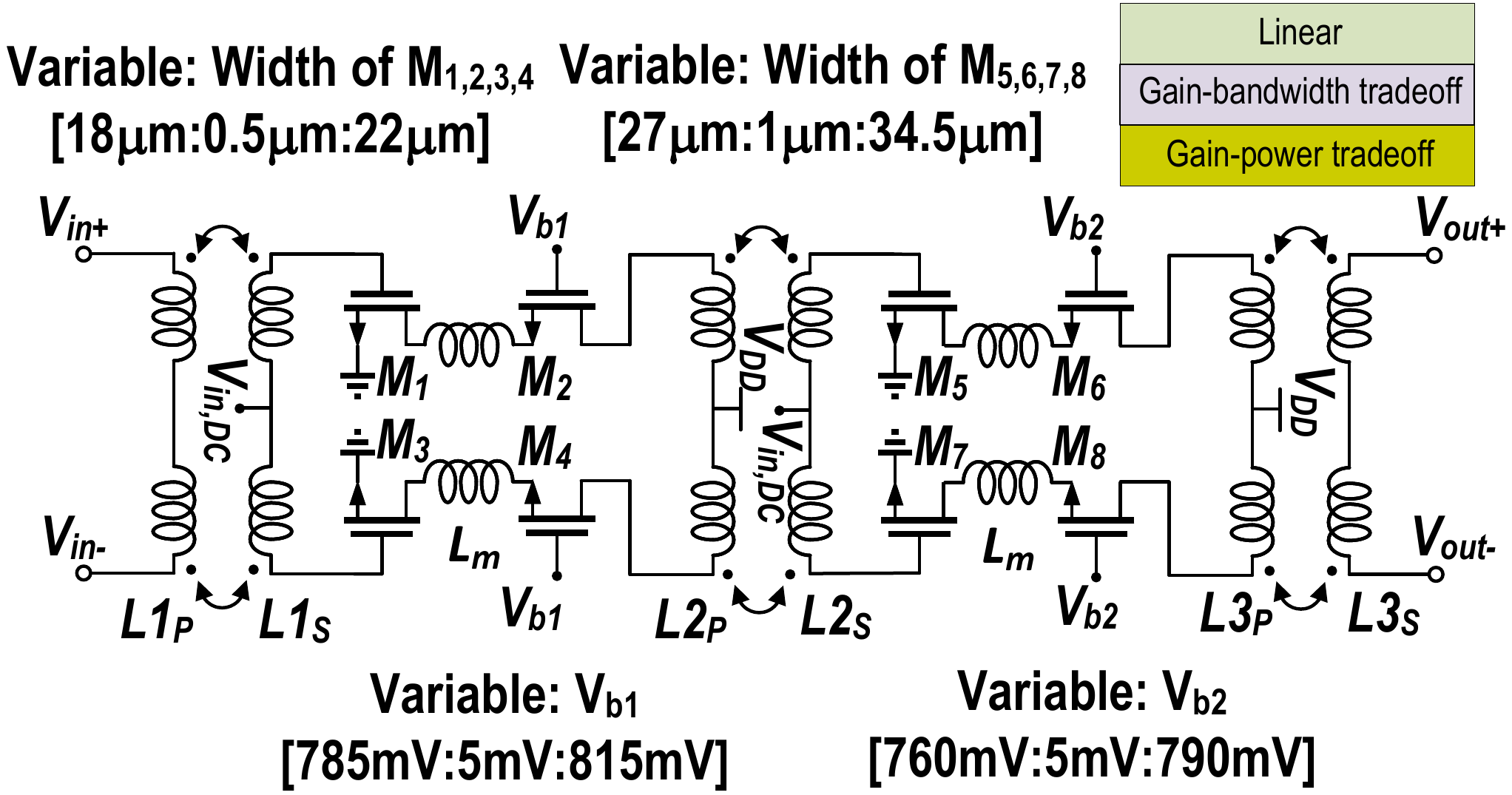}
    \caption{Schematic of a power amplifier (PA).}
    \label{circuitpa}
\end{figure}

A wireless communication transmitter requires a power amplifier to amplify the transmitted signal and deliver more power to the antenna, in order to mitigate the propagation loss of the electromagnetic waves and cover a longer operational distance \cite{niknejad}. An efficient design of a two-stage differential cascode amplifier \cite{PAref} that can provide sufficient power gain, while showing efficiency in terms of power consumption, may depend on multiple design parameters (\cref{circuitpa}(a)).

% \yc{$f(R^3) \rightarrow  R^7$ }
%\bc {\color{brown} \st{The second non linear non radio frequency circuits we tested is Power Amplifier, it shows a} 
Similar to LNA, nearly all data filtering methods performed well for the PA circuit, although with higher cross-run variance, with the exception of the baseline $D^m_\epsilon$ and the ablation $\bar{D}^m_\epsilon$ (\cref{res:pa}).
We conclude that both LNA and PA — highly non-linear circuits with intricate tradeoffs, whose design has never before been automated — are easily learned within the operational range tested in this work.
%\rc{Compared to the LNA, we observe a high variance across different runs}.
%We hypothesize that although Power Amplifier is similar to the LNA in that the performance can be easily met, some of the \st{special circuit parameter property} \rc{what is this term?} make it really sensitive to data splitting. We can especially observe this by comparing data size plot \rc{versus something else!?} where 5\% training data \st{have} ]rc{leads} to a huge variance.

%\begin{figure}[!hbt]
%    \centering
%          \includegraphics[width=0.8\columnwidth]{fig/p2/pa/subset-0.9-margin.png}%
%          \label{fig:left}%
%    \caption{PA, comparing datasets}
%    \label{PA1}
%%    \vspace{-0.15in}
%\end{figure}

% Inductance of inductors used as transoformators, Width of Transistors in both stages, and $V_b$ are some of the parameters studied in for this topology. Besides, Power Added Efficiency, Power Gain, and Drain Efficiency have been selected  as figures of merit for this Circuit. 

% \begin{figure}[hpt]
%     \centering
%     \includegraphics[width=0.5\columnwidth]{LNA.JPG}
%     \caption{LNA}
%     \label{LNA}
% \end{figure}

\subsubsection{Mixer}

An essential component in frequency conversion of modern radio-frequency and millimeter-wave circuits is a mixer.
Given two input signals at frequencies $f_1$ and $f_2$, a mixer %(depending on the type of operation) 
can generate desired signals at subtraction and summation frequencies, i.e., $f_{\Delta}=|f_1-f_2|$ and $f_{\Sigma}=f_1+f_2$. Shown in \cref{circuit_LNA_Mixer}(b) is a common schematic for mixers known as a Gilbert Cell \cite{1049925}. This circuit operates by having radio frequency (RF) and local oscillation (LO) signals as the inputs and multiplying them to generate a signal with the summation or subtraction frequencies.

% \yc{$f(R^3) \rightarrow  R^4$ }
The mixer is a sufficiently complex circuit that different data filtering methods achieve different performance when learning to design it (\cref{res:mixer}).
%By applying different methods on mixer circuit, we can clearly see 
Our proposed method, $\bar{D}^*_\epsilon$, achieves near-perfect success even at very low error margins, and only our other method, $\bar{D}^*_0$, matches it at 5\% error.
Interestingly, the naïve perturbed dataset $D_\epsilon$ also performs much better than the baseline and ablation methods.
%data modification method outperforms other methods, the BestEpsilonFeasible method almost reaches 100\% test success rate at 0.5\% - 1\% evaluation margin while the average for other methods is around 50\% at that evaluation margin.}

%\begin{figure}[!hbt]
%    \centering
%          \includegraphics[width=0.8\columnwidth]{fig/p2/mixer/subset-0.9-margin.png}%
%          \label{fig:left}%
%    \caption{Mixer, comparing datasets}
%    \label{Mixer1}
%%    \vspace{-0.15in}
%\end{figure}

% The combined design parameters in this circuits include Load Resistance, Width of differential transistors, Width of tail transistors and, LO signal DC Level. Performance metrics considered in study of this circuit are Harmonic Distortion, RF to If gain, and noise figure.

%Mixers, serve an invaluable role of frequency conversion in the communication circuits. Transmitter and receiver employ these circuits respectively as upconversion circuits and downconversion circuits. These circuits, due to operating with different frequencies as their inputs and output, show rather nonlinear behaviour, and hence, are challenging to design. In this work a downconversion mixer for IEEE802.11a is considered as shown in fig\ref{Mixer}. \rf{importance of mixer}

% \begin{figure}[hpt]
%     \centering
%     \includegraphics[width=0.5\columnwidth]{Mixer.JPG}
%     \caption{Gilbert Cell Mixer}
%     \label{Mixer}
% \end{figure}

\subsubsection{Voltage-Controlled Oscillators (VCO)}
\begin{figure}
    \centering
    \includegraphics[width=0.75\columnwidth]{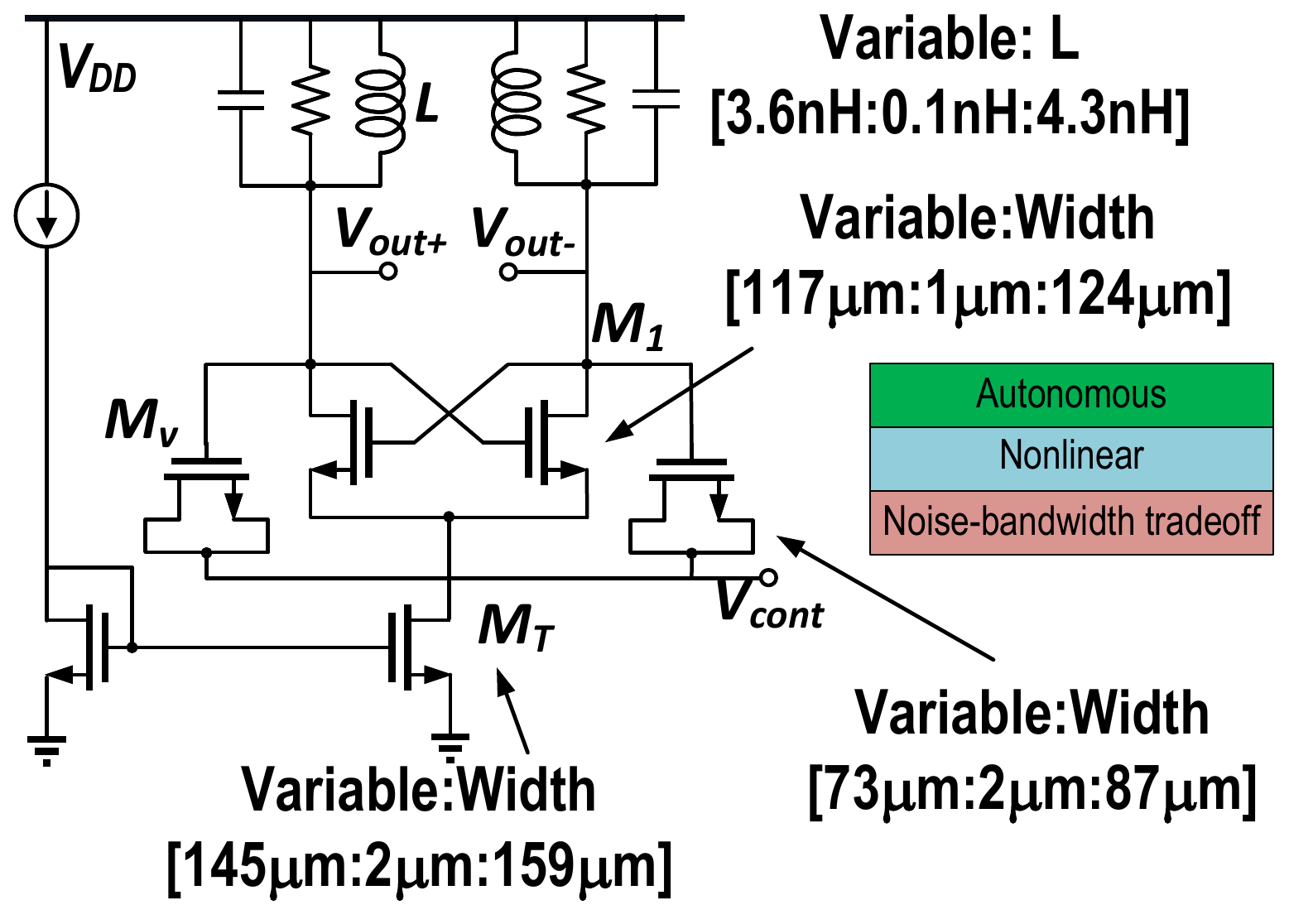}
    \caption{Schematic of a voltage-controlled oscillator (VCO).}
    \label{vco}
\end{figure}

A critical circuit block in RF applications is the oscillator, specifically voltage-controlled versions with frequency tuning capability \cite{VCO}, which is responsible for generating a sustainable periodic output autonomously.
Shown in \cref{vco} is a CMOS cross-coupled VCO \cite{alihaj}.
VCO's desired behavior is to vary output frequency within a required tuning range with control voltage variation \cite{razavi}.
The transistors consume DC power to compensate for any physical losses while the electromagnetic energy exchange among the capacitors and the inductors leads to a sustainable oscillation.

% \yc{$f(R^3) \rightarrow  R^4$ }

Automatically designing a VCO circuit proved a challenging task for all of the tested method (\cref{res:vco}).
%By testing various methods on the VCO, \rc{as shown in Fig. \ref{Vco1}} we can clearly see that 
The order of relative performance was similar to the mixer, with our proposed method, $\bar{D}^*_\epsilon$ outperforming the others with 83.5\% success rate at 5\% error margin.
Achieving near-perfect success on this circuit therefore remains an open challenge for future research.
%data modification method outperforms other methods similar to the mixer circuit; the \st{BestEpsilonFeasible} \rc{abbreviation} method reaches 80\% test success rate at 0.5\% - 1\% evaluation margin while the average for other methods is around 50\% at the same evaluation margin.

\subsection{Clustering Effect Analysis}
Since our approach involves the replacement of circuit parameters with alternative parameters within the parameter space that yield improved performance, fewer distinct parameters remain after filtering than initially simulated. In situations where multiple performance metrics are mapped to the same parameter vector, it becomes intriguing to investigate the potential impact of this clustering on the performance of our algorithm. After constructing our dataset, we count the number of distinct circuits in the constructed dataset as well as the perplexity of the resulting dataset. The resulting perplexity is defined as $PP(p)=2^{H(p)}$ where entropy is estimated over the constructed dataset based on the counts of the distinct parameters.
In the majority of circuits, the number of distinct parameter vectors in the constructed dataset is only approximately 3-12\% of its size. In the majority of circuits, the number of distinct parameter vectors in the constructed dataset is only approximately 3-12\% of its size. Specifically, for the Cascode circuit, we find 489 distinct circuits, which are 12\% of the dataset size, while the resulting dataset's perplexity is 286. In the case of more advanced circuits like Mixer and VCO, we observe a distinct count of 2-3\% of the data size. The resulting perplexity is 45 for Mixer and 68 for VCO, both amounting to roughly 1\% of their respective dataset sizes. The evident clustering effect observed in our construction method holds significant implications. One plausible hypothesis we put forward is that this clustering phenomenon could potentially be advantageous for training Machine Learning Models. By leveraging this clustering effect, we can effectively sidestep issues arising from overlapping mappings, enabling us to construct sample-efficient dataset.

%However, even in extreme cases, such as CS Amplifier with an unchanged rate of nearly 50\% of the original parameters, we observe perplexity of 8913, which is 36\% of the constructed dataset. Our algorithm consistently outperforms alternative methods. Hence, clustering does not significantly affect the performance of our algorithm.

% Since our approach involves the replacement of circuit parameters with alternative parameters within the parameter space that yield improved performance, fewer distinct parameters remain after filtering than originally simulated. In situations where multiple performance metrics are mapped to the same parameters vector, it becomes intriguing to investigate the potential impact of this clustering on the performance of our algorithm. We count the number of unique remaining parameters  after constructing our dataset. In the majority of circuits, only approximately 5-10\% of the original data points remain unchanged. However, even in extreme cases, such as NMOS circuits with an unchanged rate of nearly 50\%, or in the case of PA circuits with an unchanged rate of almost 0.01\%, our algorithm consistently outperforms alternative methods. Hence, we conclude that clustering does not significantly affect the performance of our algorithm. 

\subsection{Performance Metric Ordering Variations}
In our study, we subjected the LNA circuit to an assessment using two distinct orders of performance metrics: Order A (Power Gain, $S_{11}$, NF), and Order B ($S_{11}$, NF, Power Gain). We optimized for maximizing Power Gain and minimizing $S_{11}$ and NF in both orders. Notably, the circuits generated by Order A showcased an average Power Gain that was larger (thus better) by 0.84 dB compared to those generated by Order B. Additionally, these circuits exhibited an average $S_{11}$ that was higher (thus worse) by 0.53 dB in comparison. We conducted a similar analysis for the Common Source Amplifier, Cascode Amplifier, and Two-Stage circuits. By prioritizing the order of the bandwidth during dataset construction, we observed circuits with higher average bandwidth. Similarly, with regards to power consumption, which we aimed to minimize, assigning the highest priority to power consumption led to the production of circuits with lower power consumption. We conclude that the user-specified order of performance metrics effectively creates the desired preference over them.

% \begin{figure}[t]
% \centering
% \hfill

% \subfigure[]{\label{res:pa}\includegraphics[width=0.48\columnwidth]{fig/p2/pa/subset-0.9-margin.png}} \hfill
% \subfigure[]{\label{res:vco}\includegraphics[width=0.48\columnwidth]{fig/p2/vco/subset-0.9-margin.png}} \hfill

% \caption{Comparing success rate in the threshold specification problem for different training datasets in two circuits: (a) power amplifier (PA); (b) voltage-controlled oscillator (VCO).\label{res:others}}
% \end{figure}

% \begin{figure}[t]
% \centering
% \hfill
% \subfigure[]{\label{res:cascode1}\includegraphics[width=0.48\columnwidth]{fig/p1_base/cascode/margin.png}} \hfill
% \subfigure[]{\label{res:cascode2}\includegraphics[width=0.48\columnwidth]{fig/p2_datasize/cascode/margin.png}}
% \hfill
% \caption{Comparing success rate in the threshold specification problem for different training datasets in two circuits: (a) power amplifier; (b) voltage-controlled oscillator.\label{res:others}}
% \end{figure}

%\begin{figure}[!hbt]
%    \centering
%          \includegraphics[width=0.8\columnwidth]{fig/p2/vco/subset-0.9-margin.png}%
%          \label{fig:left}%
%    \caption{VCO, comparing datasets}
%    \label{Vco1}
%%    \vspace{-0.15in}
%\end{figure}

% The input parameters are the width of top current device M2, PMOS pair M0, M1 and Varactors. Performance Metrics are Tuning range, Phase noise and Power Consumption.  
% \begin{figure}[hpt]
%     \centering
%     \includegraphics[width=0.8\columnwidth]{PA.JPG}
%     \caption{Power Amplifier}
%     \label{PA}
% \end{figure}

\section{Conclusion}
We present a data filtering pipeline that can generate, from a circuit simulation dataset, a training dataset for supervised learning of a circuit design agent for threshold specification.
%generation method to \rc{modified or modify} circuit data to meet threshold specification. 
In extensive experiments with several baselines on a variety of linear, nonlinear, and autonomous analog and radio-frequency circuits, we find that our proposed method performs near-perfectly in all but the hardest circuit.
This supports our hypothesis that a systematic selection of representative circuits can alleviate the “non-injective” property of the simulator function, which is vastly exacerbated by the threshold specification setting.
%helped machine learning model to learn the relationship between circuit performance specification and circuit parameters easier.
%We demonstrate the robustness of our method on 7 different topologies of \rc{linear, nonlinear, and autonomous radio-frequency and mm-wave} circuits.  
%\yc{Our proposed data generation method achieves 95 percent accuracy or higher given the margin of 1 percent for all the circuits except the VCO circuit. For the VCO circuit our method produces the highest score of 83 percent accuracy given the margin of 1 percent. } 
The results also show the sample efficiency of our method.
While not directly comparable with previous work, we often use a number of simulations an order of magnitude or more smaller than ever before, and learning from even 5\% of this data is often highly successful as well.
% by testing how data size variations will affect performance. For most \rc{of the examined} circuits, even using 5\% of the training data can achieve acceptable performance within circuit \st{generation} \rc{design} industry standard.
Lastly, we also show that our method is, to some extent, model agnostic by training with different machine learning methods and comparing their performance. To the best of our knowledge, this is the first time that a wide collection of analog circuits at various frequencies and of varied operations have been extensively examined and shown capable of being automatically designed.
% while the proposed method outperforms the prior art.} 
We believe that the methods and results of this work can help the growth of the circuit design industry by addressing the rapidly increasing demand for advanced electronic chipsets.

\section*{Acknowledgements}
Roy Fox is partly supported by the Hasso Plattner Foundation.

\bibliography{iclr2023_conference}

\begin{thebibliography}{42}
\providecommand{\natexlab}[1]{#1}
\providecommand{\url}[1]{\texttt{#1}}
\expandafter\ifx\csname urlstyle\endcsname\relax
  \providecommand{\doi}[1]{doi: #1}\else
  \providecommand{\doi}{doi: \begingroup \urlstyle{rm}\Url}\fi

\bibitem[Abbasi et~al.(2010)Abbasi, Kjellberg, de~Graauw, van~der Heijden,
  Roovers, and Zirath]{PAref}
Abbasi, M., Kjellberg, T., de~Graauw, A., van~der Heijden, E., Roovers, R., and
  Zirath, H.
\newblock A broadband differential cascode power amplifier in 45 nm cmos for
  high-speed 60 ghz system-on-chip.
\newblock In \emph{2010 IEEE Radio Frequency Integrated Circuits Symposium},
  pp.\  533--536. IEEE, 2010.

\bibitem[Afacan et~al.(2021)Afacan, Louren{\c{c}}o, Martins, and
  D{\"u}ndar]{survey2}
Afacan, E., Louren{\c{c}}o, N., Martins, R., and D{\"u}ndar, G.
\newblock Machine learning techniques in analog/rf integrated circuit design,
  synthesis, layout, and test.
\newblock \emph{Integration}, 77:\penalty0 113--130, 2021.

\bibitem[Bandler \& Chen(1988)Bandler and Chen]{Bandler1988}
Bandler, J. and Chen, S.
\newblock Circuit optimization: the state of the art.
\newblock \emph{IEEE Transactions on Microwave Theory and Techniques},
  36\penalty0 (2):\penalty0 424--443, 1988.
\newblock \doi{10.1109/22.3532}.

\bibitem[Bandler \& Rayas-Sánchez(2023)Bandler and
  Rayas-Sánchez]{Bandler2023}
Bandler, J.~W. and Rayas-Sánchez, J.~E.
\newblock An early history of optimization technology for automated design of
  microwave circuits.
\newblock \emph{IEEE Journal of Microwaves}, 3\penalty0 (1):\penalty0 319--337,
  2023.
\newblock \doi{10.1109/JMW.2022.3225012}.

\bibitem[Boyd et~al.(2005)Boyd, Kim, Patil, and Horowitz]{optim1}
Boyd, S.~P., Kim, S.-J., Patil, D.~D., and Horowitz, M.~A.
\newblock Digital circuit optimization via geometric programming.
\newblock \emph{Operations research}, 53\penalty0 (6):\penalty0 899--932, 2005.

\bibitem[Breiman(2001)]{breiman2001random}
Breiman, L.
\newblock Random forests.
\newblock \emph{Machine learning}, 45\penalty0 (1):\penalty0 5--32, 2001.

\bibitem[Brunvand(2010)]{digitalcad2}
Brunvand, E.
\newblock \emph{Digital VLSI chip design with Cadence and Synopsys CAD tools}.
\newblock Addison-Wesley New York, 2010.

\bibitem[Dai \& Harjani(2003)Dai and Harjani]{VCO}
Dai, L. and Harjani, R.
\newblock \emph{Design of high-performance CMOS voltage-controlled
  oscillators}.
\newblock Springer Science \& Business Media, 2003.

\bibitem[Devi et~al.(2021)Devi, Tilwankar, and Zele]{Devi2021}
Devi, S., Tilwankar, G., and Zele, R.
\newblock Automated design of analog circuits using machine learning
  techniques.
\newblock In \emph{2021 25th International Symposium on VLSI Design and Test
  (VDAT)}, pp.\  1--6, 2021.
\newblock \doi{10.1109/VDAT53777.2021.9601131}.

\bibitem[Dumesnil et~al.(2014)Dumesnil, Nabki, and Boukadoum]{Dumesnil2014}
Dumesnil, E., Nabki, F., and Boukadoum, M.
\newblock Rf-lna circuit synthesis by genetic algorithm-specified artificial
  neural network.
\newblock In \emph{2014 21st IEEE International Conference on Electronics,
  Circuits and Systems (ICECS)}, pp.\  758--761, 2014.
\newblock \doi{10.1109/ICECS.2014.7050096}.

\bibitem[Fukuda et~al.(2017{\natexlab{a}})Fukuda, Ishii, and Takai]{Fukuda2017}
Fukuda, M., Ishii, T., and Takai, N.
\newblock Op-amp sizing by inference of element values using machine learning.
\newblock In \emph{2017 International Symposium on Intelligent Signal
  Processing and Communication Systems (ISPACS)}, pp.\  622--627,
  2017{\natexlab{a}}.
\newblock \doi{10.1109/ISPACS.2017.8266553}.

\bibitem[Fukuda et~al.(2017{\natexlab{b}})Fukuda, Ishii, and Takai]{datasize_1}
Fukuda, M., Ishii, T., and Takai, N.
\newblock Op-amp sizing by inference of element values using machine learning.
\newblock In \emph{2017 International Symposium on Intelligent Signal
  Processing and Communication Systems (ISPACS)}, pp.\  622--627,
  2017{\natexlab{b}}.
\newblock \doi{10.1109/ISPACS.2017.8266553}.

\bibitem[Gilbert(1968)]{1049925}
Gilbert, B.
\newblock A precise four-quadrant multiplier with subnanosecond response.
\newblock \emph{IEEE Journal of Solid-State Circuits}, 3\penalty0 (4):\penalty0
  365--373, 1968.
\newblock \doi{10.1109/JSSC.1968.1049925}.

\bibitem[Gray \& Meyer(1982)Gray and Meyer]{meyer}
Gray, P.~R. and Meyer, R.~G.
\newblock Mos operational amplifier design-a tutorial overview.
\newblock \emph{Ieee journal of solid-state circuits}, 17\penalty0
  (6):\penalty0 969--982, 1982.

\bibitem[Grover \& Chaudhary(2014)Grover and Chaudhary]{optim3}
Grover, G. and Chaudhary, I.
\newblock Implementation of particle swarm optimization algorithm in vhdl for
  digital circuits optimization.
\newblock \emph{International Journal of Information Engineering and Electronic
  Business}, 6\penalty0 (5):\penalty0 16, 2014.

\bibitem[Hajimiri \& Lee(1999)Hajimiri and Lee]{alihaj}
Hajimiri, A. and Lee, T.~H.
\newblock Design issues in cmos differential lc oscillators.
\newblock \emph{IEEE Journal of Solid-State Circuits}, 34\penalty0
  (5):\penalty0 717--724, 1999.

\bibitem[Harsha \& Harish(2020)Harsha and Harish]{2020artificial}
Harsha, M. and Harish, B.
\newblock Artificial neural network model for design optimization of 2-stage
  op-amp.
\newblock In \emph{2020 24th International Symposium on VLSI Design and Test
  (VDAT)}, pp.\  1--5. IEEE, 2020.

\bibitem[Hassan et~al.(2016)Hassan, Mohamed, Rabie, and Etman]{hassan2016novel}
Hassan, A.-K.~S., Mohamed, A.~S., Rabie, A.~A., and Etman, A.~S.
\newblock A novel surrogate-based approach for optimal design of
  electromagnetic-based circuits.
\newblock \emph{Engineering Optimization}, 48\penalty0 (2):\penalty0 185--198,
  2016.

\bibitem[Kamal(2022)]{kamal}
Kamal, K.~Y.
\newblock The silicon age: Trends in semiconductor devices industry.
\newblock \emph{Journal of Engineering Science \& Technology Review},
  15\penalty0 (1), 2022.

\bibitem[Kingma \& Ba(2015)Kingma and Ba]{adam}
Kingma, D.~P. and Ba, J.
\newblock Adam: A method for stochastic optimization.
\newblock In \emph{ICLR (Poster)}, 2015.
\newblock URL \url{http://arxiv.org/abs/1412.6980}.

\bibitem[Ko \& Lin(2006)Ko and Lin]{cascode}
Ko, S. and Lin, J.
\newblock A linearized cascode cmos power amplifier.
\newblock In \emph{2006 IEEE Annual Wireless and Microwave Technology
  Conference}, pp.\  1--4. IEEE, 2006.

\bibitem[Kouhalvandi et~al.(2018)Kouhalvandi, Ceylan, and Yagci]{keysight}
Kouhalvandi, L., Ceylan, O., and Yagci, H.~B.
\newblock Power amplifier design optimization with simultaneous cooperation of
  eda tool and numeric analyzer.
\newblock In \emph{2018 18th Mediterranean Microwave Symposium (MMS)}, pp.\
  202--205. IEEE, 2018.

\bibitem[Kunz \& Pradhan(1994)Kunz and Pradhan]{optim2}
Kunz, W. and Pradhan, D.~K.
\newblock Recursive learning: a new implication technique for efficient
  solutions to cad problems-test, verification, and optimization.
\newblock \emph{IEEE Transactions on Computer-Aided Design of Integrated
  Circuits and Systems}, 13\penalty0 (9):\penalty0 1143--1158, 1994.

\bibitem[Lannutti et~al.(2012)Lannutti, Nenzi, and Olivieri]{netlist}
Lannutti, F., Nenzi, P., and Olivieri, M.
\newblock Klu sparse direct linear solver implementation into ngspice.
\newblock In \emph{Proceedings of the 19th International Conference Mixed
  Design of Integrated Circuits and Systems-MIXDES 2012}, pp.\  69--73. IEEE,
  2012.

\bibitem[Lerdworatawee \& Namgoong(2005)Lerdworatawee and Namgoong]{LNASD}
Lerdworatawee, J. and Namgoong, W.
\newblock Wide-band cmos cascode low-noise amplifier design based on source
  degeneration topology.
\newblock \emph{IEEE Transactions on Circuits and Systems I: Regular Papers},
  52\penalty0 (11):\penalty0 2327--2334, 2005.

\bibitem[Louren{\c{c}}o et~al.(2018)Louren{\c{c}}o, Rosa, Martins, Aidos,
  Canelas, P{\'o}voa, and Horta]{2018exploration}
Louren{\c{c}}o, N., Rosa, J., Martins, R., Aidos, H., Canelas, A., P{\'o}voa,
  R., and Horta, N.
\newblock On the exploration of promising analog ic designs via artificial
  neural networks.
\newblock In \emph{2018 15th International Conference on Synthesis, Modeling,
  Analysis and Simulation Methods and Applications to Circuit Design (SMACD)},
  pp.\  133--136. IEEE, 2018.

\bibitem[Lourenço et~al.(2018)Lourenço, Rosa, Martins, Aidos, Canelas,
  Póvoa, and Horta]{datasize_3}
Lourenço, N., Rosa, J., Martins, R., Aidos, H., Canelas, A., Póvoa, R., and
  Horta, N.
\newblock On the exploration of promising analog ic designs via artificial
  neural networks.
\newblock In \emph{2018 15th International Conference on Synthesis, Modeling,
  Analysis and Simulation Methods and Applications to Circuit Design (SMACD)},
  pp.\  133--136, 2018.
\newblock \doi{10.1109/SMACD.2018.8434896}.

\bibitem[Lourenço et~al.(2019)Lourenço, Afacan, Martins, Passos, Canelas,
  Póvoa, Horta, and Dundar]{Lourenco2019}
Lourenço, N., Afacan, E., Martins, R., Passos, F., Canelas, A., Póvoa, R.,
  Horta, N., and Dundar, G.
\newblock Using polynomial regression and artificial neural networks for
  reusable analog ic sizing.
\newblock In \emph{2019 16th International Conference on Synthesis, Modeling,
  Analysis and Simulation Methods and Applications to Circuit Design (SMACD)},
  pp.\  13--16, 2019.
\newblock \doi{10.1109/SMACD.2019.8795282}.

\bibitem[Martin(2002)]{cadence}
Martin, A. J.~L.
\newblock Cadence design environment.
\newblock \emph{New Mexico State University, Tutorial paper}, pp.\ ~35, 2002.

\bibitem[Micheli(1994)]{digitalcad1}
Micheli, G.~D.
\newblock \emph{Synthesis and optimization of digital circuits}.
\newblock McGraw-Hill Higher Education, 1994.

\bibitem[Mina et~al.(2022)Mina, Jabbour, and Sakr]{mina2022}
Mina, R., Jabbour, C., and Sakr, G.~E.
\newblock A review of machine learning techniques in analog integrated circuit
  design automation.
\newblock \emph{Electronics}, 11\penalty0 (3):\penalty0 435, 2022.

\bibitem[Murphy \& McCarthy(2021)Murphy and McCarthy]{2021automated}
Murphy, S.~D. and McCarthy, K.~G.
\newblock Automated design of cmos operational amplifier using a neural
  network.
\newblock In \emph{2021 32nd Irish Signals and Systems Conference (ISSC)}, pp.\
   1--6. IEEE, 2021.

\bibitem[M.V. \& Harish(2020{\natexlab{a}})M.V. and Harish]{HarshaHarish}
M.V., H. and Harish, B.~P.
\newblock Artificial neural network model for design optimization of 2-stage
  op-amp.
\newblock In \emph{2020 24th International Symposium on VLSI Design and Test
  (VDAT)}, pp.\  1--5, 2020{\natexlab{a}}.
\newblock \doi{10.1109/VDAT50263.2020.9190315}.

\bibitem[M.V. \& Harish(2020{\natexlab{b}})M.V. and Harish]{datasize_4}
M.V., H. and Harish, B.~P.
\newblock Artificial neural network model for design optimization of 2-stage
  op-amp.
\newblock In \emph{2020 24th International Symposium on VLSI Design and Test
  (VDAT)}, pp.\  1--5, 2020{\natexlab{b}}.
\newblock \doi{10.1109/VDAT50263.2020.9190315}.

\bibitem[Nenzi \& Vogt(2011)Nenzi and Vogt]{ngspice}
Nenzi, P. and Vogt, H.
\newblock Ngspice users manual version 23, 2011.

\bibitem[Niknejad et~al.(2012)Niknejad, Chowdhury, and Chen]{niknejad}
Niknejad, A.~M., Chowdhury, D., and Chen, J.
\newblock Design of cmos power amplifiers.
\newblock \emph{IEEE Transactions on Microwave Theory and Techniques},
  60\penalty0 (6):\penalty0 1784--1796, 2012.

\bibitem[Razavi(2012)]{razavi}
Razavi, B.
\newblock \emph{RF microelectronics}, volume~2.
\newblock Prentice hall New York, 2012.

\bibitem[Renner \& Ek{\'a}rt(2003)Renner and Ek{\'a}rt]{genetic}
Renner, G. and Ek{\'a}rt, A.
\newblock Genetic algorithms in computer aided design.
\newblock \emph{Computer-aided design}, 35\penalty0 (8):\penalty0 709--726,
  2003.

\bibitem[Settaluri et~al.(2020)Settaluri, Haj-Ali, Huang, Hakhamaneshi, and
  Nikolic]{2020autockt}
Settaluri, K., Haj-Ali, A., Huang, Q., Hakhamaneshi, K., and Nikolic, B.
\newblock Autockt: Deep reinforcement learning of analog circuit designs.
\newblock In \emph{2020 Design, Automation \& Test in Europe Conference \&
  Exhibition (DATE)}, pp.\  490--495. IEEE, 2020.

\bibitem[Vural et~al.(2015)Vural, Kahraman, Erkmen, and Yildirim]{2015process}
Vural, R., Kahraman, N., Erkmen, B., and Yildirim, T.
\newblock Process independent automated sizing methodology for current steering
  dac.
\newblock \emph{International Journal of Electronics}, 102\penalty0
  (10):\penalty0 1713--1734, 2015.

\bibitem[Wang et~al.(2018{\natexlab{a}})Wang, Yang, Lee, and Han]{2018learning}
Wang, H., Yang, J., Lee, H.-S., and Han, S.
\newblock Learning to design circuits.
\newblock \emph{arXiv preprint arXiv:1812.02734}, 2018{\natexlab{a}}.

\bibitem[Wang et~al.(2018{\natexlab{b}})Wang, Luo, and Gong]{datasize_2}
Wang, Z., Luo, X., and Gong, Z.
\newblock Application of deep learning in analog circuit sizing.
\newblock In \emph{Proceedings of the 2018 2nd International Conference on
  Computer Science and Artificial Intelligence}, CSAI '18, pp.\  571–575, New
  York, NY, USA, 2018{\natexlab{b}}. Association for Computing Machinery.
\newblock ISBN 9781450366069.
\newblock \doi{10.1145/3297156.3297160}.
\newblock URL \url{https://doi.org/10.1145/3297156.3297160}.

\end{thebibliography}
\bibliographystyle{iclr2023_conference}

\appendix
\clearpage

\onecolumn
\section{Appendix} 

\subsection{Model Architecture}
\label{appendix_a} 

In this work, we propose a Multi-layer perceptron (MLP) architecture with seven layers for the task at hand. The first layer takes in a vector of size equal to the number of performance metrics for the circuit and outputs a vector of length 200. The last layer takes in a vector of length 200 and outputs the same number of parameters as in the circuit. The middle five layers are constant across all circuits and have the following [input, output] size configurations: [200, 300], [300, 500], [500, 500], [500, 300], [300, 200]. Each layer is separated by the Rectified Linear Unit (ReLU) activation function. We trained each MLP model for 100 epochs using the Adam optimizer \cite{adam} with a default learning rate of 0.001. Additionally, we also trained a Random Forest (RF) model with the default number of trees (100) and default arguments.

\subsection{Comparing Methods}
\label{appendix_b} 

\begin{figure*}[!hb]
\centering
\hfill
\subfigure[]{\includegraphics[width=0.31\textwidth]{fig/p2_methods/cascode/subset-0.9-margin.png}} \hfill
\subfigure[]{\includegraphics[width=0.31\textwidth]{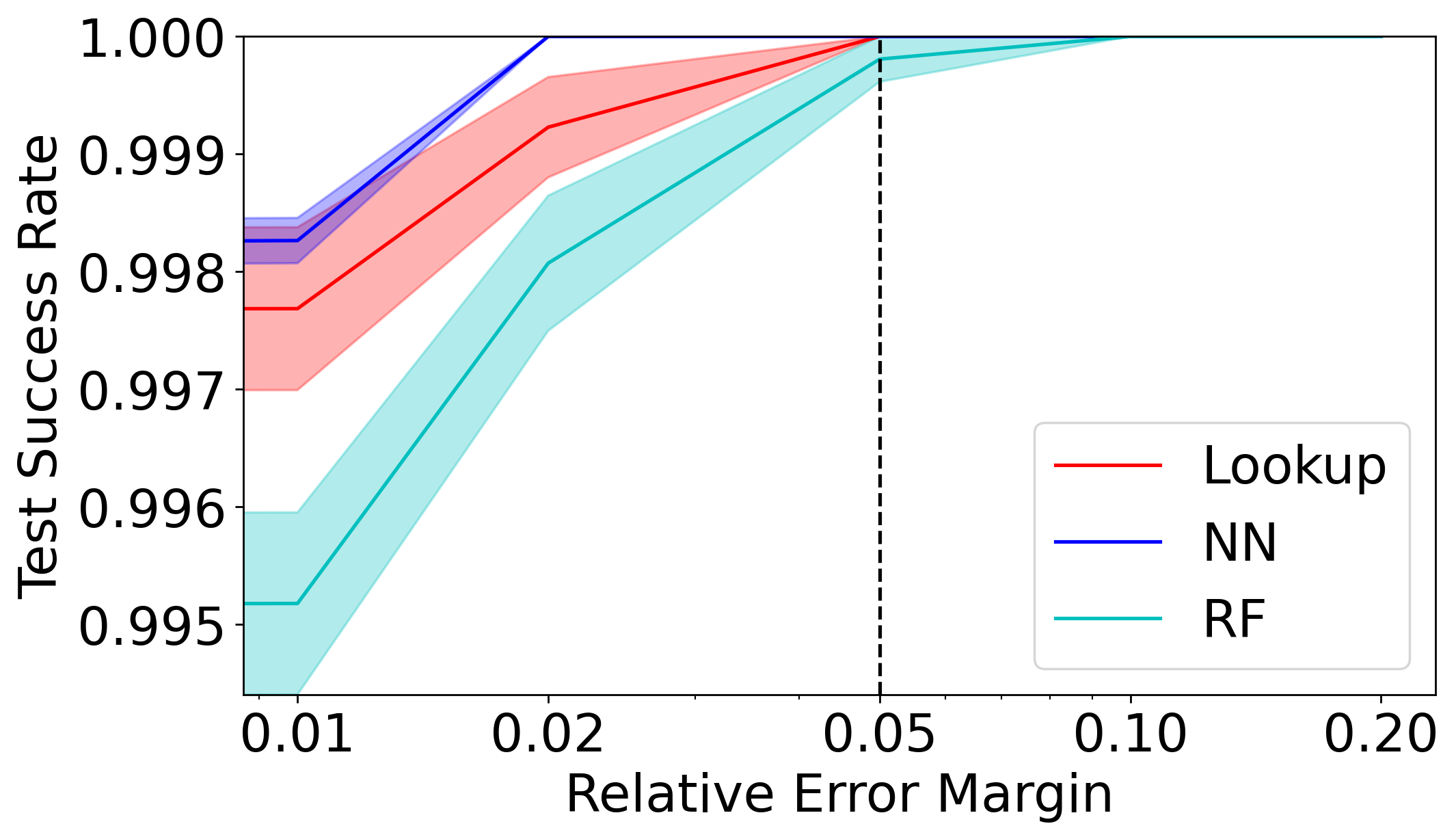}} \hfill
\subfigure[]{\includegraphics[width=0.31\textwidth]{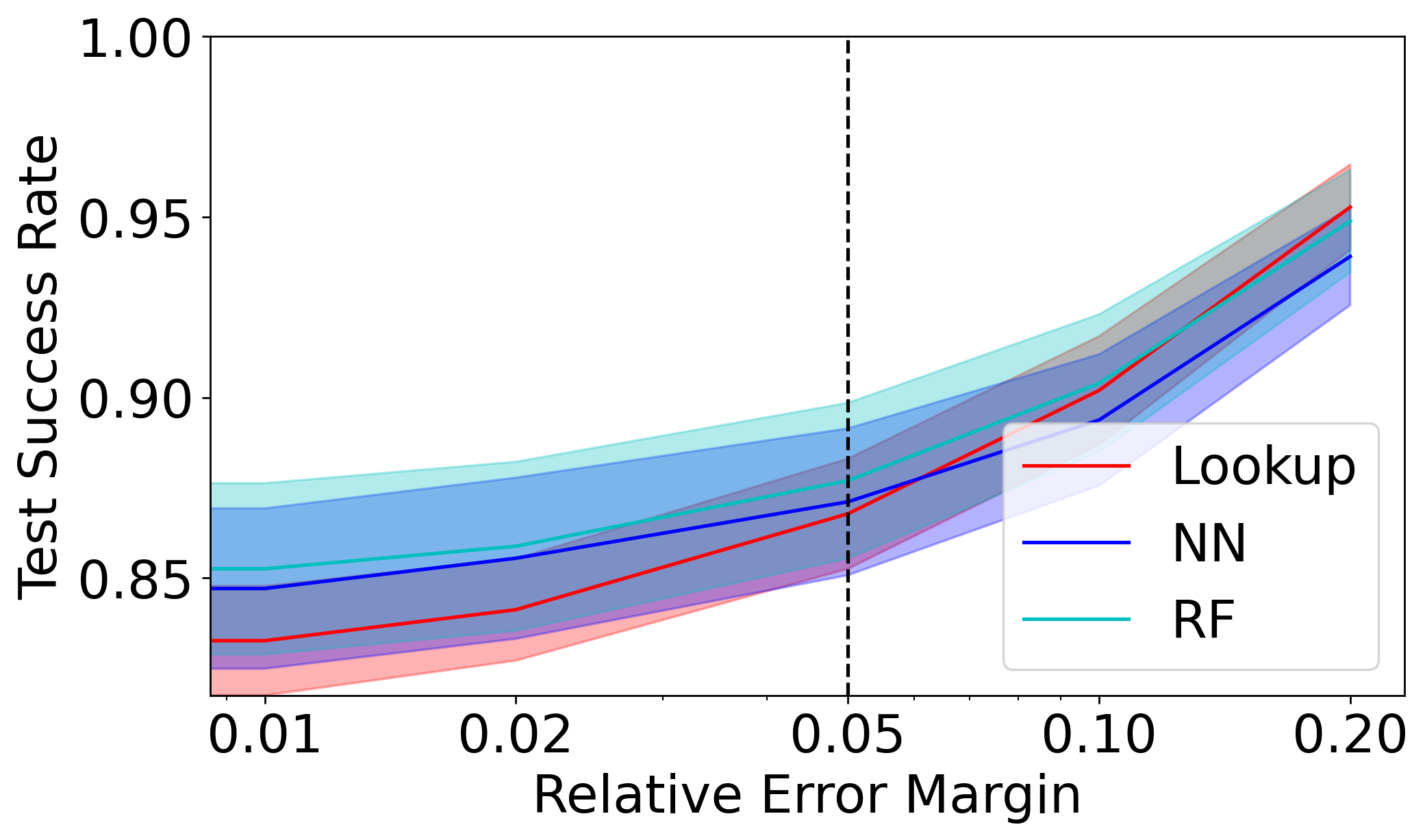}} \hfill

\caption{Comparing different ML methods in three circuits: (a) cascode; (b) low-noise amplifier; (c) voltage-controlled oscillator.\label{res:appendix_methods}}
\end{figure*}

% \begin{figure}[!hbt]
%     \centering
%           \includegraphics[width=0.8\columnwidth]{fig/p2_methods/two_stage/subset-0.9-margin.png}%
%           \label{fig:left}%
%     \caption{TwoStage, comparing methods}
%     \label{p2_two_stage3}
% \end{figure}

\begin{figure*}[!hb]
\centering
\hfill
\subfigure[]{\includegraphics[width=0.31\textwidth]{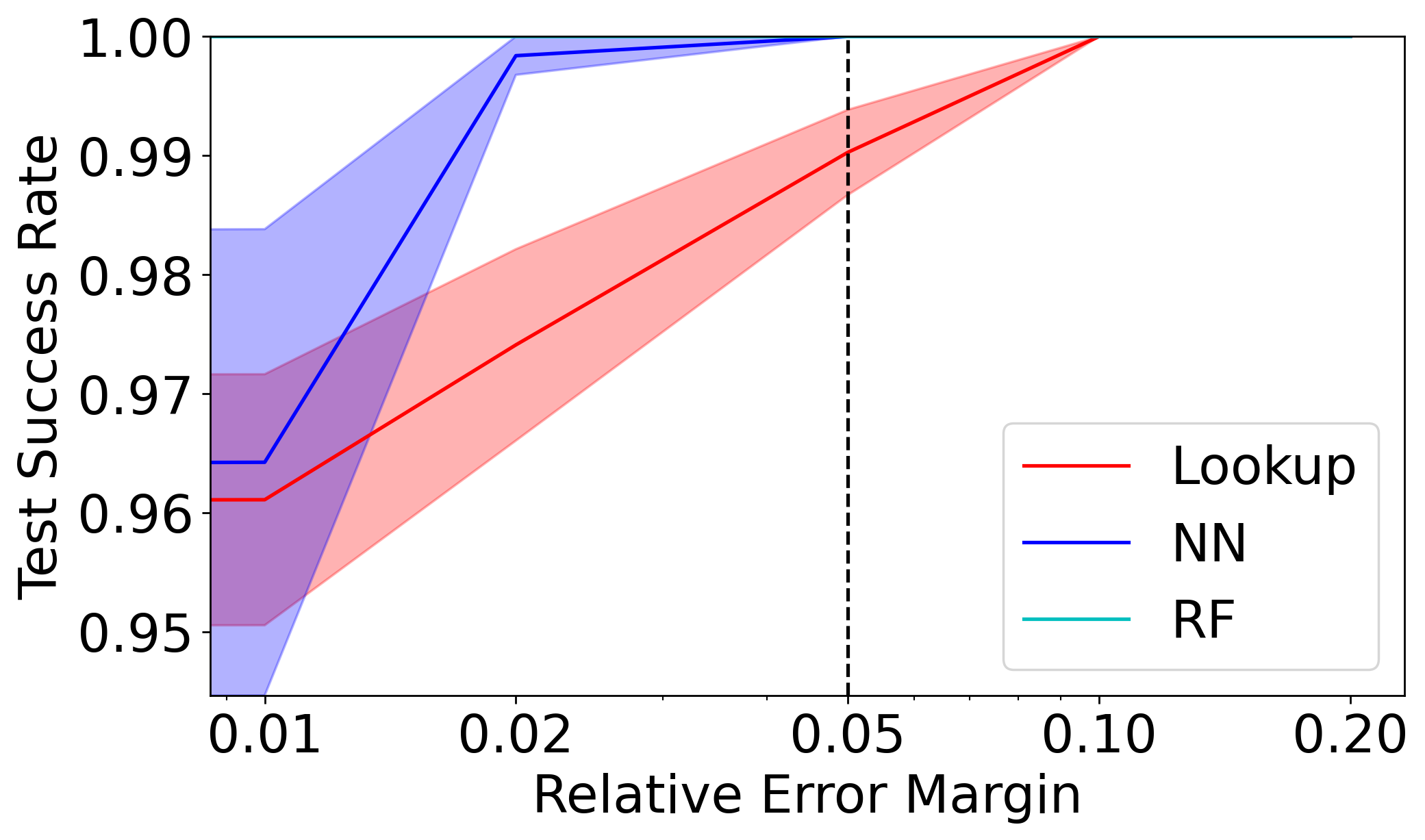}} \hfill
\subfigure[]{\includegraphics[width=0.31\textwidth]{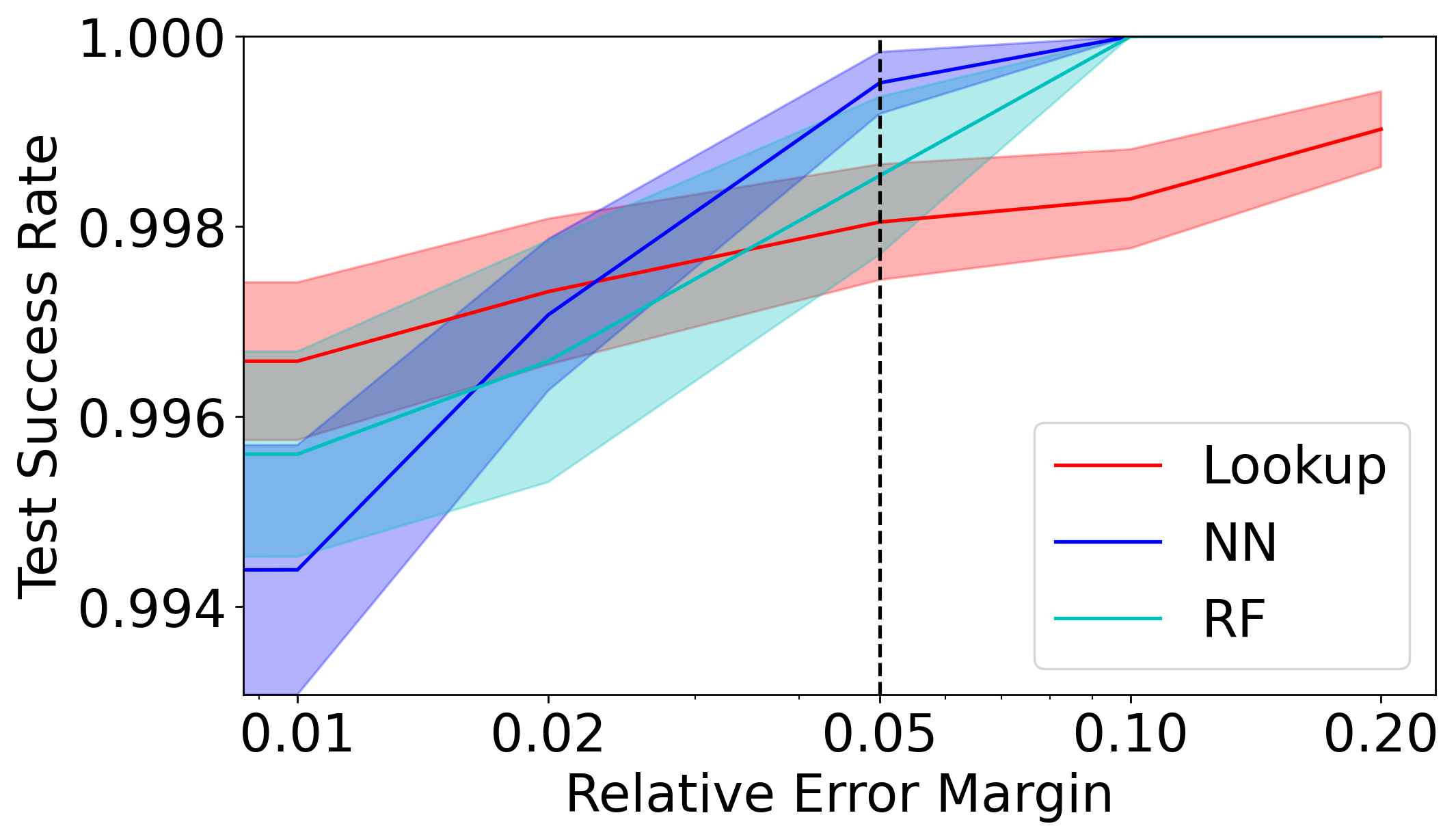}} \hfill
\subfigure[]{\includegraphics[width=0.31\textwidth]{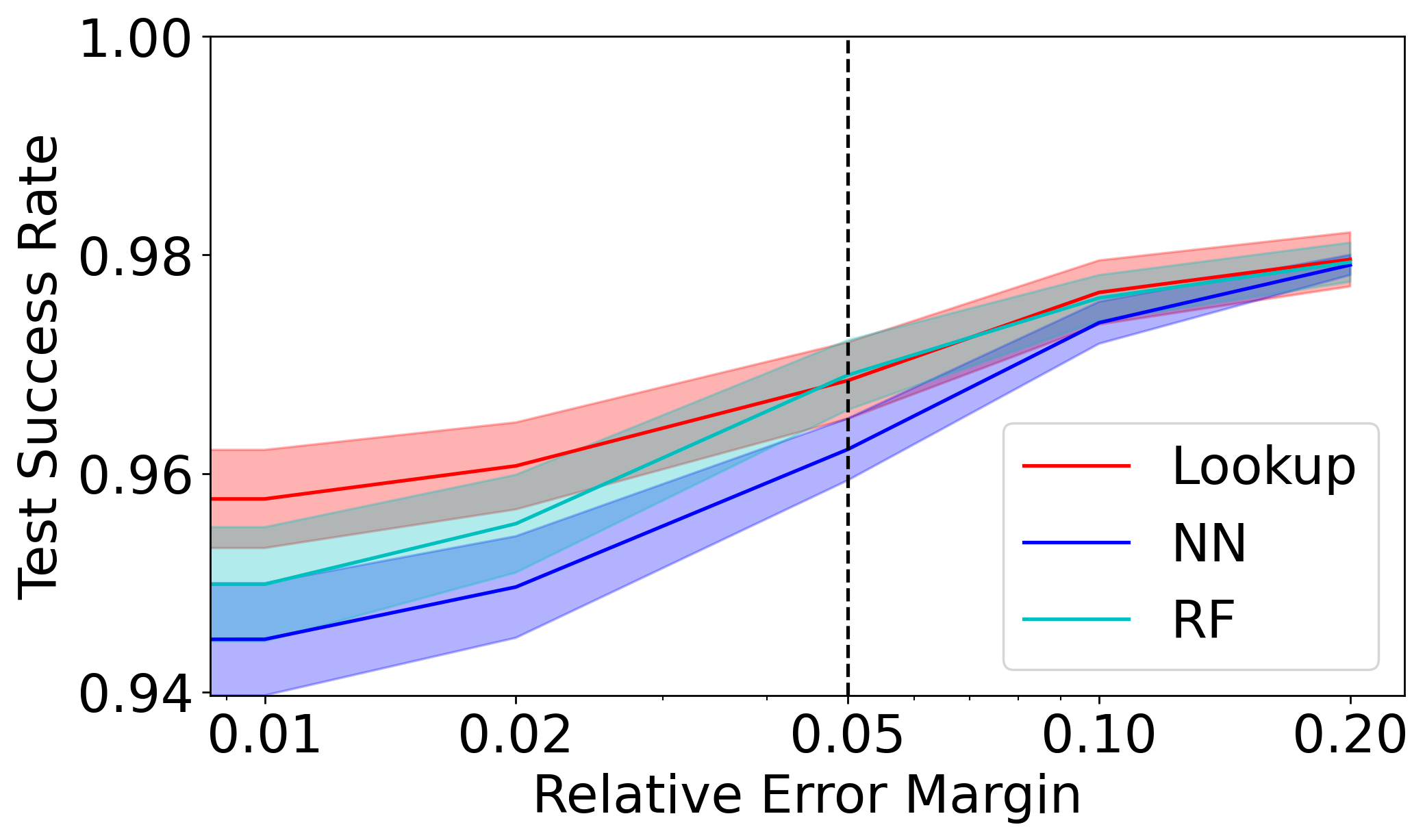}}
\hfill
\caption{Comparing different ML methods in three circuits:  (a) two stage; (b) mixer; and (c)  power amplifier.\label{res:appendix_methods}}
\end{figure*}

We compared the performance of three different machine learning methods: Neural Network (NN), Random Forest (RF), and Lookup Table (LT) using a ten-fold cross-validation setup with 90 \% of the data used for training.
The purpose of this work was to demonstrate that our method is model-agnostic, and thus, we did not attempt to fine-tune the NN or RF models. The lookup table approach was implemented by searching for the circuit with the lowest error in the training dataset for a given testing circuit.
The NN and RF models were trained as specified in Section \ref{appendix_a}. The results in table \ref{table_method} suggest that all three methods perform similarly, with most of the circuits achieving 95-100\% accuracy and a margin of 1\%. This suggests that even a simple model like RF can produce good results. The detailed plots are presented in \ref{res:appendix_methods}.

\begin{table*}[!hbt]
 \small
% \singlespacing
 \centering
\caption{Circuit Method Comparison at 1 $\%$ }\label{table_method}
\begin{tabular}{|p{1.4cm}|p{1.8cm}|p{1.8cm}|p{1.8cm}|p{1.8cm}|p{1.8cm}|p{1.8cm}|p{1.8cm}|}
\hline
\textbf{ML/Circuit} & \textbf{CS} & \textbf{Cascode} &\textbf{Two Stage} & \textbf{LNA}  &\textbf{PA}   &\textbf{Mixer} &\textbf{VCO} \\
\hline
Lookup &0.94 $\pm$ 0.005 &0.888 $\pm$ 0.005 &0.961 $\pm$ 0.011 &0.998 $\pm$ 0.001 &$\textbf{0.958}$ $\pm$ 0.004 &0.997 $\pm$ 0.001 &0.833 $\pm$ 0.015\\ 
 \hline 
NN &0.854 $\pm$ 0.023 & $\textbf{0.971}$ $\pm$ 0.004 &$\textbf{0.964}$ $\pm$ 0.02 &$\textbf{0.998}$ $\pm$ 0.0 &0.945 $\pm$ 0.005 &0.994 $\pm$ 0.001 &0.847 $\pm$ 0.022\\ 
 \hline 
RF & $\textbf{0.953}$ $\pm$ 0.004 &0.944 $\pm$ 0.005 &1.0 $\pm$ 0.0 &0.995 $\pm$ 0.001 &0.95 $\pm$ 0.005 &0.996 $\pm$ 0.001 &$\textbf{0.853}$ $\pm$ 0.024\\ 
 \hline

\end{tabular}
\end{table*}

%%%%Methods Graphs%%%%%%%

% \begin{figure}[!hbt]
%     \centering
%           \includegraphics[width=0.3\columnwidth]{fig/p2_methods/cascode/subset-0.9-margin.png}%
%           \label{fig:left}%
%     \caption{Cascode, comparing methods}
%     \label{p2_cascode}
% \end{figure}

% \begin{figure}[!hbt]
%     \centering
%           \includegraphics[width=0.3\columnwidth]{fig/p2_methods/lna/subset-0.9-margin.png}%
%           \label{fig:left}%
%     \caption{LNA, comparing methods}
%     \label{LNA3}
% \end{figure}

% \begin{figure}[!hbt]
%     \centering
%           \includegraphics[width=0.3\columnwidth]{fig/p2_methods/vco/subset-0.9-margin.png}%
%           \label{fig:left}%
%     \caption{VCO, comparing methods}
%     \label{Vco3}
% \end{figure}

% \begin{figure}[!hbt]
%     \centering
%           \includegraphics[width=0.3\columnwidth]{fig/p2_methods/mixer/subset-0.9-margin.png}%
%           \label{fig:left}%
%     \caption{Mixer, comparing methods}
%     \label{Mixer3}
% \end{figure}

% \begin{figure}[!hbt]
%     \centering
%           \includegraphics[width=0.3\columnwidth]{fig/p2_methods/pa/subset-0.9-margin.png}%
%           \label{fig:left}%
%     \caption{PA, comparing methods}
%     \label{PA3}
% \end{figure}
%%%%

\label{appendix_c} 
\subsection{Comparing Datasizes}

\begin{table*}[!hbt]
 \small
% \singlespacing
 \centering
\caption{Circuit Data Size Comparison for $D_0$ at 1 $\%$ }\label{table1}
\begin{tabular}{|p{1.4cm}|p{1.8cm}|p{1.8cm}|p{1.8cm}|p{1.8cm}|p{1.8cm}|p{1.8cm}|p{1.8cm}|}
\hline
\textbf{\%/Circuit} & \textbf{CS} & \textbf{Cascode} &\textbf{Two Stage} & \textbf{LNA}  &\textbf{PA}   &\textbf{Mixer} &\textbf{VCO} \\
\hline
0.05 &0.295 $\pm$ 0.036 &0.207 $\pm$ 0.029 &0.566 $\pm$ 0.04 &0.944 $\pm$ 0.005 &0.712 $\pm$ 0.026 &0.319 $\pm$ 0.037 &0.375 $\pm$ 0.033\\ 
 \hline 
0.1 &0.278 $\pm$ 0.041 &0.13 $\pm$ 0.025 &0.676 $\pm$ 0.077 &0.933 $\pm$ 0.012 &0.711 $\pm$ 0.02 &0.282 $\pm$ 0.055 &0.342 $\pm$ 0.049\\ 
 \hline 
0.2 &0.351 $\pm$ 0.047 &0.292 $\pm$ 0.067 &0.719 $\pm$ 0.078 &0.991 $\pm$ 0.003 &0.759 $\pm$ 0.029 &0.364 $\pm$ 0.05 &0.38 $\pm$ 0.089\\ 
 \hline 
0.5 &0.54 $\pm$ 0.053 &0.492 $\pm$ 0.065 &0.813 $\pm$ 0.106 &0.992 $\pm$ 0.007 &0.689 $\pm$ 0.091 &0.387 $\pm$ 0.021 &0.452 $\pm$ 0.011\\ 
 \hline 
0.9 &0.539 $\pm$ 0.051 &0.583 $\pm$ 0.053 &0.927 $\pm$ 0.022 &0.998 $\pm$ 0.001 &0.73 $\pm$ 0.014 &0.396 $\pm$ 0.032 &0.454 $\pm$ 0.055\\ 
 \hline 

\end{tabular}
\end{table*}

\begin{table*}[!hbt]
 \small
% \singlespacing
 \centering
\caption{Circuit Data Size Comparison for $\bar{D}^{*}_{\epsilon}$ at 1 $\%$ }\label{table1}
\begin{tabular}{|p{1.4cm}|p{1.8cm}|p{1.8cm}|p{1.8cm}|p{1.8cm}|p{1.8cm}|p{1.8cm}|p{1.8cm}|}
\hline
\textbf{\%/Circuit} & \textbf{CS} & \textbf{Cascode} &\textbf{Two Stage} & \textbf{LNA}  &\textbf{PA}   &\textbf{Mixer} &\textbf{VCO} \\
\hline
0.05 &0.83 $\pm$ 0.018 &0.803 $\pm$ 0.019 &0.817 $\pm$ 0.021 &0.972 $\pm$ 0.003 &0.874 $\pm$ 0.026 &0.927 $\pm$ 0.01 &0.819 $\pm$ 0.009\\ 
 \hline 
0.1 &0.751 $\pm$ 0.035 &0.815 $\pm$ 0.029 &0.962 $\pm$ 0.015 &0.985 $\pm$ 0.003 &0.927 $\pm$ 0.01 &0.942 $\pm$ 0.013 &0.82 $\pm$ 0.013\\ 
 \hline 
0.2 &0.849 $\pm$ 0.029 &0.886 $\pm$ 0.019 &0.978 $\pm$ 0.004 &0.991 $\pm$ 0.002 &0.944 $\pm$ 0.013 &0.969 $\pm$ 0.017 &0.838 $\pm$ 0.019\\ 
 \hline 
0.5 &0.917 $\pm$ 0.027 &0.962 $\pm$ 0.009 &0.938 $\pm$ 0.0 &0.996 $\pm$ 0.003 &0.963 $\pm$ 0.002 &0.995 $\pm$ 0.002 &0.814 $\pm$ 0.017\\ 
 \hline 
0.9 &0.847 $\pm$ 0.04 &0.971 $\pm$ 0.004 &0.973 $\pm$ 0.018 &0.997 $\pm$ 0.001 &0.945 $\pm$ 0.006 &0.995 $\pm$ 0.001 &0.787 $\pm$ 0.037\\ 
 \hline

\end{tabular}
\end{table*}

\begin{figure*}[!hbt]
\centering
\hfill
\subfigure[]{\label{res:22s}\includegraphics[width=0.24\textwidth]{fig/p1_base/cascode/margin.png}} \hfill
\subfigure[]{\label{res:2lna}\includegraphics[width=0.24\textwidth]{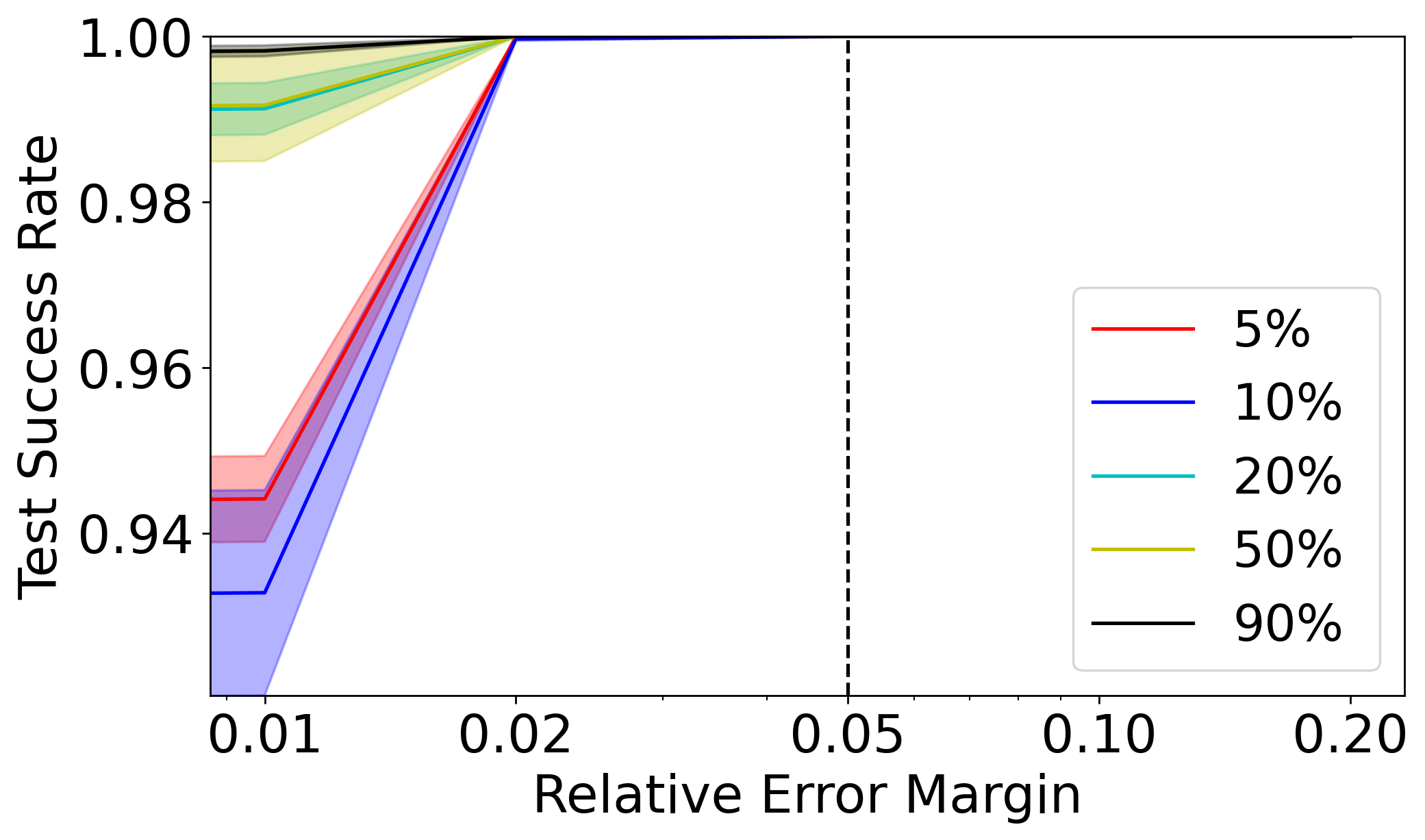}} \hfill
\subfigure[]{\label{res:2pa}\includegraphics[width=0.24\textwidth]{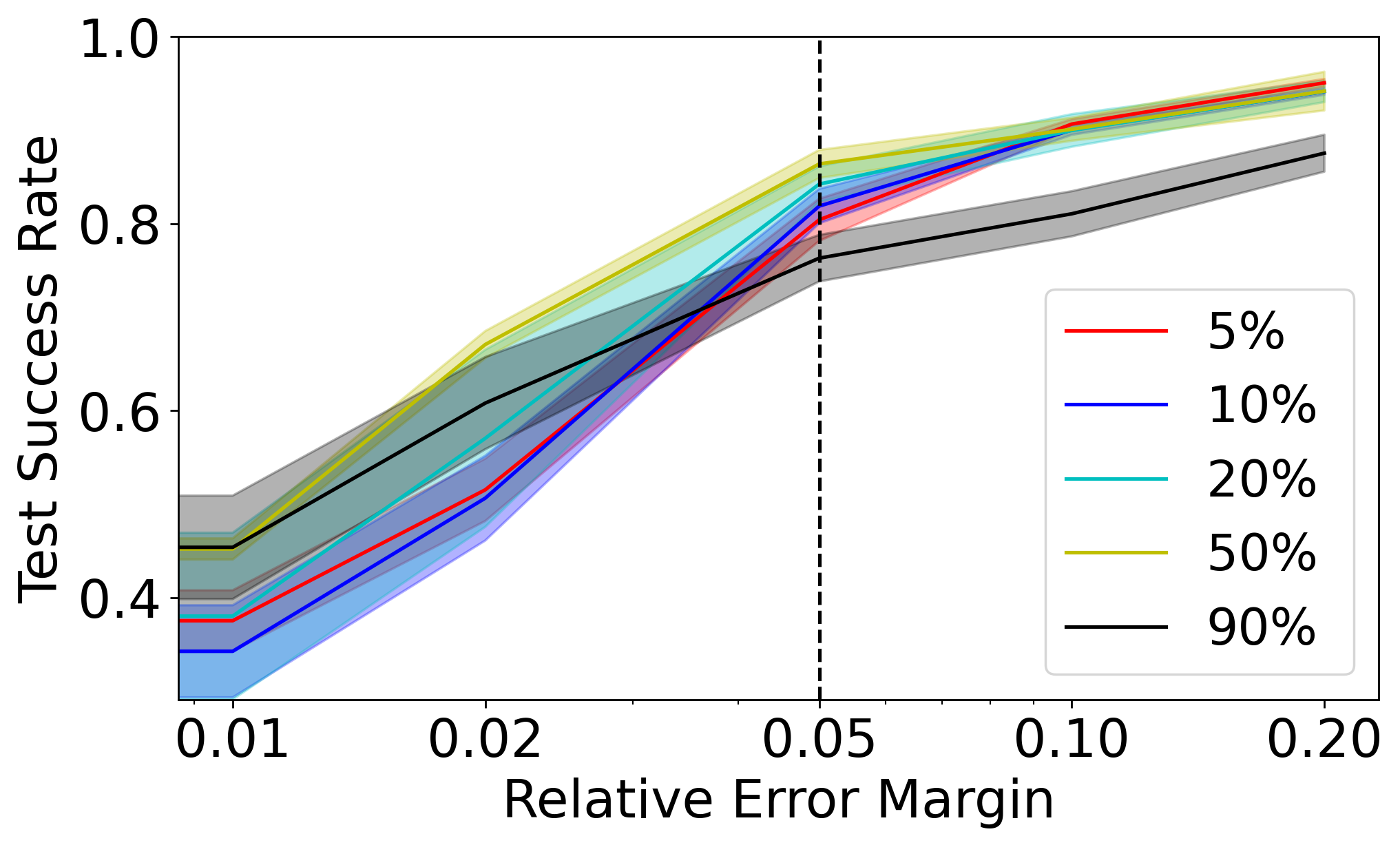}} \hfill
\subfigure[]{\label{res:2pa}\includegraphics[width=0.24\textwidth]{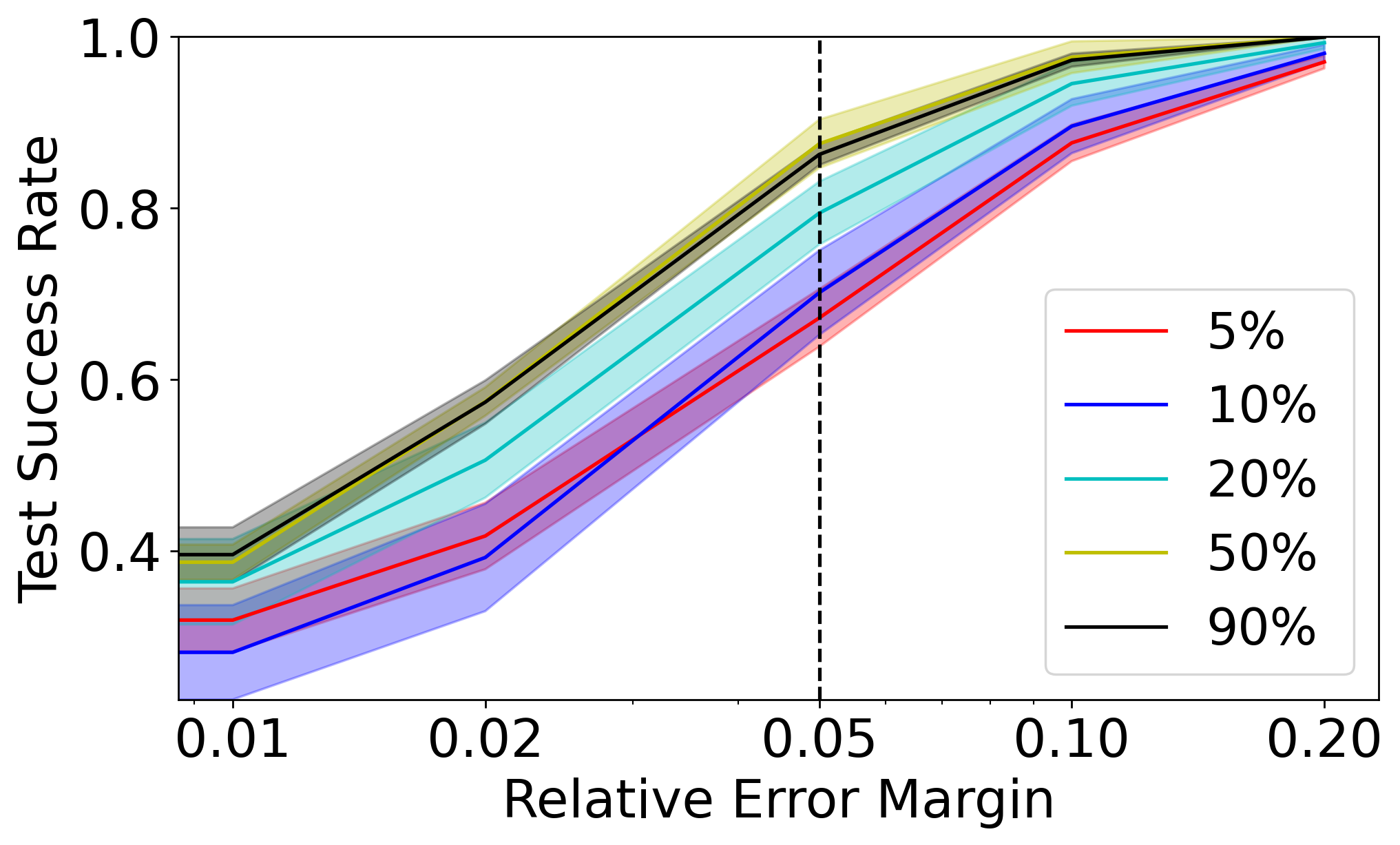}} \hfill
\caption{Comparing different datasizes for $D_0$ in circuits: (a) cascode; (b) low-noise amplifier (LNA); (c) voltage-controlled oscillator (VCO); (d) mixer.\label{res:appendix_ds}}
\end{figure*}

\begin{figure*}[!hbt]
\centering
\hfill
\subfigure[]{\label{res:2nmos}\includegraphics[width=0.24\textwidth]{fig/p1_base/nmos/margin.png}} \hfill
\subfigure[]{\label{res:2pa}\includegraphics[width=0.24\textwidth]{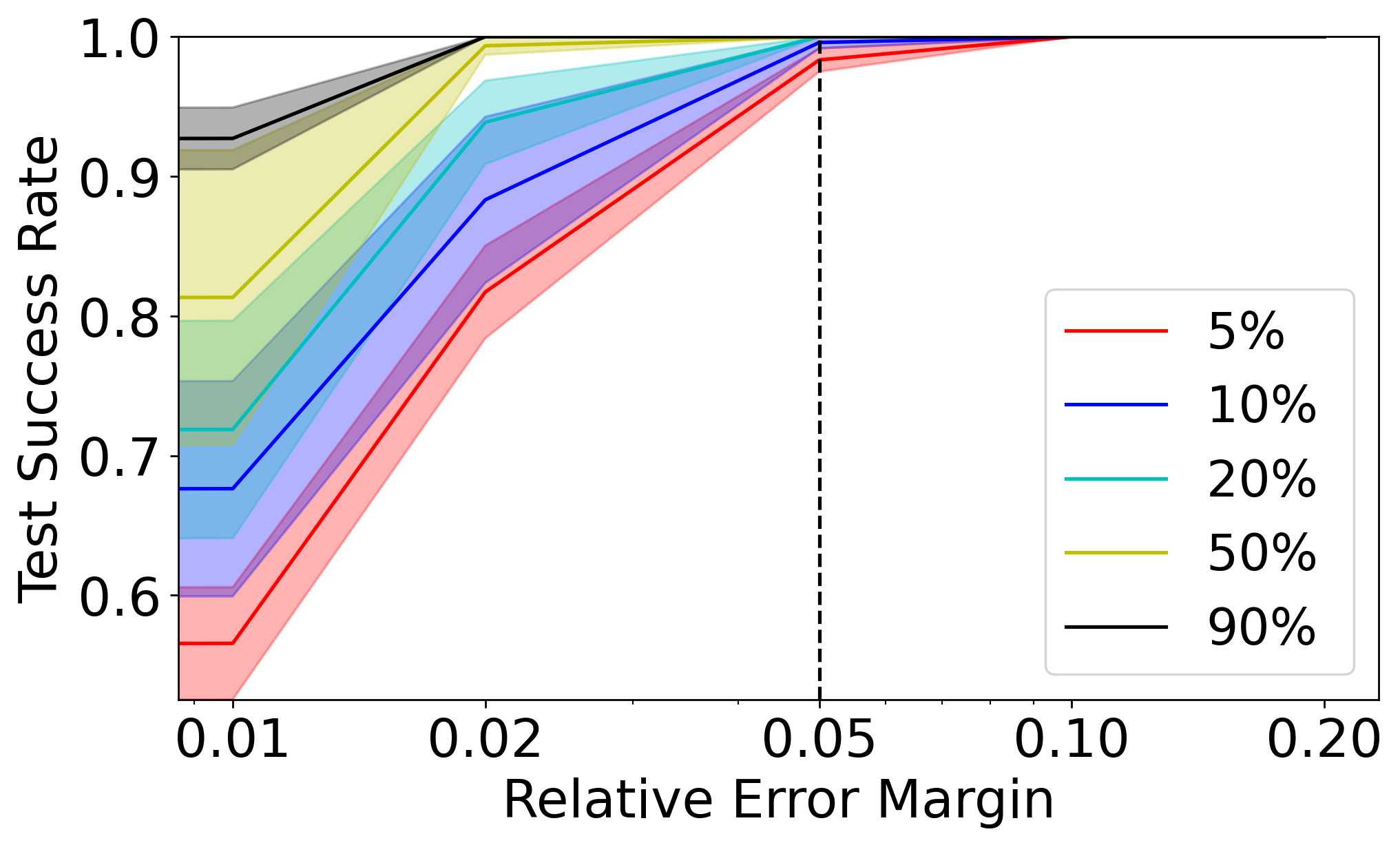}} \hfill
\subfigure[]{\label{res:p2a}\includegraphics[width=0.24\textwidth]{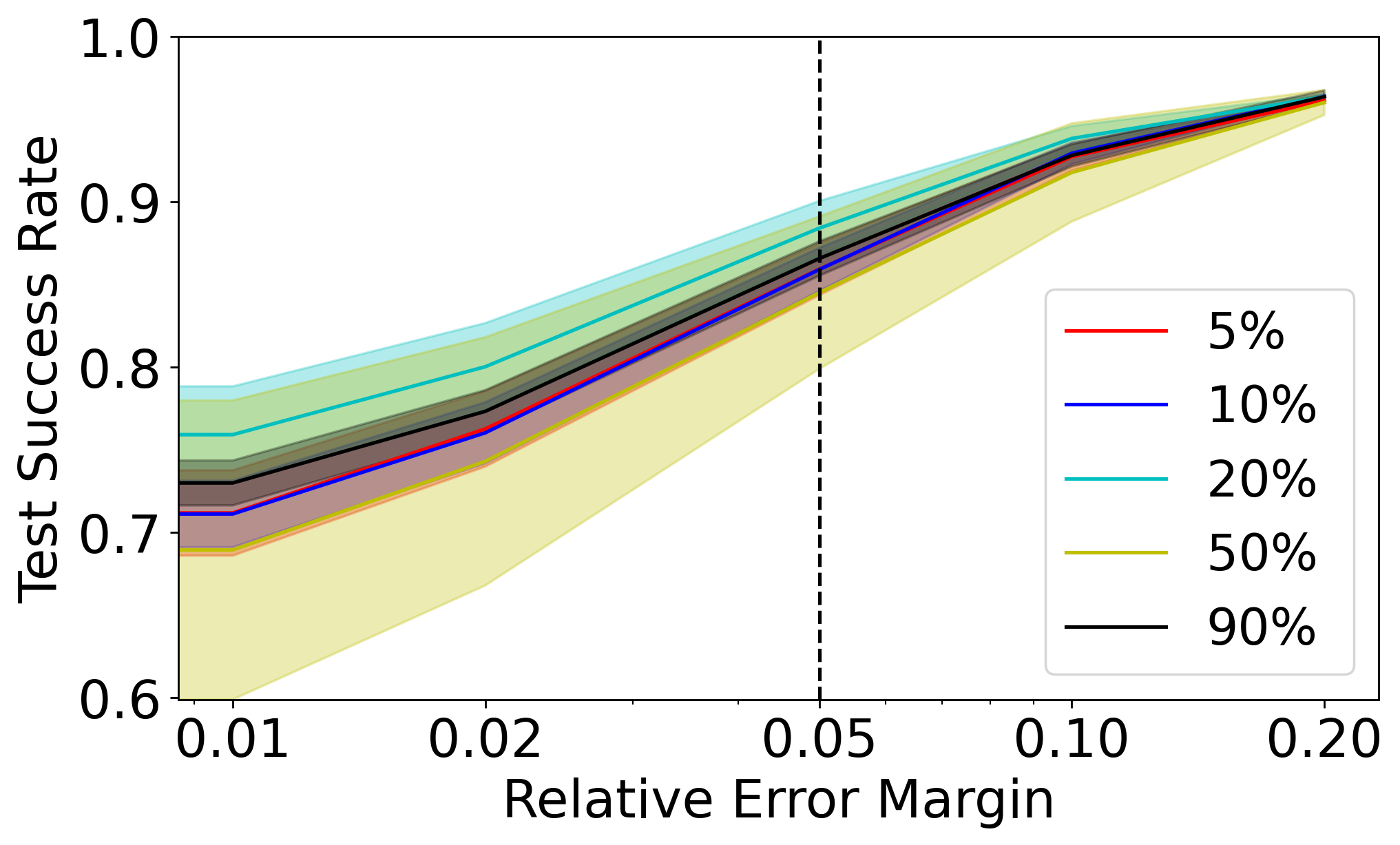}} \hfill
\caption{Comparing different datasizes for $D_0$ in circuits: (a) common source amplifier (CS) ; (b) two-stage; (c) power amplifier (PA).\label{res:appendix_ds2}}
\end{figure*}

In this study, we assess the performance of various circuits by utilizing subsamples of varying sizes, including 5\%, 10\%, 20\%, 50\%, and 90\% of the data. The 90\% subsample corresponds to a range of 2700-3600 points, while the 5\% subsample corresponds to approximately 150-200 points. Results indicate that the accuracy of the Two Stage circuit linearly increases from 56\% to 93\% as the subsample size increases from 5\% to 90\%. Similarly, for the $\bar{D_{\epsilon}^*}$ method, accuracy increases from 81\% to 97\%. Notably, all circuits exhibit an accuracy higher than 80\%, even when using only 5\% of the data. However, the accuracy of the $D_0$ circuit is observed to drop to 20\% for certain subsamples.

\label{appendix_d} 
\subsection{Comparing Datasets}

In addition to our previous results, we present a table of circuit evaluations for each method at a 1\% margin. It can be observed that the Power Amplifier, VCO, and Mixer circuits achieve the highest scores when utilizing our method. However, for less complex circuits such as the Common Source Amplifier, the $D_{\epsilon}$ method yields a 2\% higher score compared to our method.

% \begin{figure*}[t]
% \centering
% \hfill

% \subfigure[]{\label{res:pa}\includegraphics[width=0.31\textwidth]{fig/p2/pa/subset-0.9-margin.png}} 
% \subfigure[]{\label{res:vco}\includegraphics[width=0.31\textwidth]{fig/p2/vco/subset-0.9-margin.png}}

% \caption{Comparing success rate in the threshold specification problem for different training datasets in two circuits: (a) power amplifier; (b) voltage-controlled oscillator.\label{res:others}}
% \end{figure*}

\begin{table*}[!hbt]
 \small
% \singlespacing
 \centering
\caption{Circuit Comparison Info at 1 $\%$ }\label{table1}
\begin{tabular}{|p{1.4cm}|p{1.8cm}|p{1.8cm}|p{1.8cm}|p{1.8cm}|p{1.8cm}|p{1.8cm}|p{1.8cm}|}
\hline
\textbf{D/Circuit} & \textbf{CS} & \textbf{Cascode} &\textbf{Two Stage} & \textbf{LNA}  &\textbf{PA}   &\textbf{Mixer} &\textbf{VCO} \\
\hline

$D_{0}$ &0.7 $\pm$ 0.036 &0.519 $\pm$ 0.043 &0.985 $\pm$ 0.01 &0.996 $\pm$ 0.001 &0.945 $\pm$ 0.005 &0.486 $\pm$ 0.06 &0.378 $\pm$ 0.054\\ 
 \hline 
$D_{\epsilon}$ & $\textbf{0.957}$ $\pm$ 0.009 &0.921 $\pm$ 0.033 & $\textbf{0.99}$ $\pm$ 0.004 & $\textbf{0.999}$ $\pm$ 0.0 &0.97 $\pm$ 0.002 &0.824 $\pm$ 0.019 &0.713 $\pm$ 0.016\\ 
 \hline 
$D^{m}_{\epsilon}$ &0.542 $\pm$ 0.014 &0.435 $\pm$ 0.026 &0.877 $\pm$ 0.033 &0.976 $\pm$ 0.003 &0.37 $\pm$ 0.048 &0.32 $\pm$ 0.011 &0.54 $\pm$ 0.021\\ 
 \hline 
$\bar{D}^{m}_{\epsilon}$ &0.801 $\pm$ 0.038 &0.578 $\pm$ 0.033 &0.518 $\pm$ 0.048 &0.8 $\pm$ 0.024 &0.768 $\pm$ 0.092 &0.401 $\pm$ 0.022 &0.501 $\pm$ 0.015\\ 
 \hline 
$\bar{D}^{*}_0$ &0.847 $\pm$ 0.021 &0.947 $\pm$ 0.019 &0.984 $\pm$ 0.011 &0.997 $\pm$ 0.001 &0.943 $\pm$ 0.004 &0.858 $\pm$ 0.025 &0.827 $\pm$ 0.019\\ 
 \hline 
$\bar{D}^{*}_{\epsilon}$ &0.936 $\pm$ 0.012 &$\textbf{0.962}$ $\pm$ 0.007 &0.981 $\pm$ 0.012 &0.998 $\pm$ 0.001 & $\textbf{0.948}$ $\pm$ 0.005 & $\textbf{0.995}$ $\pm$ 0.001 & $\textbf{0.835}$ $\pm$ 0.016\\ 
 \hline 

\end{tabular}
\end{table*}

% \begin{table*}[!hbt]
%  \small
% % \singlespacing
%  \centering
% \caption{Circuit Data Size Comparison for $\bar{D}^{*}_{\epsilon}$ at 5 $\%$ }\label{table_5percent}
% \begin{tabular}{|p{2.0cm}|p{1.8cm}|p{1.8cm}|p{1.8cm}|p{1.8cm}|p{1.8cm}|p{1.8cm}|p{1.8cm}|}
% \hline
% \textbf{Data \% /Circuit Name} & \textbf{CS} & \textbf{Cascode} &\textbf{Two Stage} & \textbf{LNA}  &\textbf{PA}   &\textbf{Mixer} &\textbf{VCO} \\
% \hline
% 0.05 &0.968 $\pm$ 0.007 &0.959 $\pm$ 0.007 &0.998 $\pm$ 0.001 &1.0 $\pm$ 0.0 &0.907 $\pm$ 0.022 &0.979 $\pm$ 0.004 &0.856 $\pm$ 0.007\\ 
%  \hline 
% 0.1 &0.959 $\pm$ 0.008 &0.966 $\pm$ 0.013 &1.0 $\pm$ 0.0 &1.0 $\pm$ 0.0 &0.952 $\pm$ 0.006 &0.983 $\pm$ 0.005 &0.858 $\pm$ 0.015\\ 
%  \hline 
% 0.2 &0.976 $\pm$ 0.007 &0.984 $\pm$ 0.001 &1.0 $\pm$ 0.0 &1.0 $\pm$ 0.0 &0.961 $\pm$ 0.008 &0.992 $\pm$ 0.005 &0.866 $\pm$ 0.019\\ 
%  \hline 
% 0.5 &0.994 $\pm$ 0.004 &0.998 $\pm$ 0.001 &1.0 $\pm$ 0.0 &1.0 $\pm$ 0.0 &0.974 $\pm$ 0.001 &0.998 $\pm$ 0.0 &0.844 $\pm$ 0.018\\ 
%  \hline 
% 0.9 &0.98 $\pm$ 0.014 &0.999 $\pm$ 0.001 &1.0 $\pm$ 0.0 &1.0 $\pm$ 0.0 &0.963 $\pm$ 0.003 &1.0 $\pm$ 0.0 &0.815 $\pm$ 0.038\\ 
%  \hline 

% \end{tabular}
% \end{table*}

\pagebreak

\label{appendix_e} 
\subsection{Comparing Circuits}
\label{appendix_f} 

As seen in Table \ref{comparing_101}, a comparison of our proposed method is conducted against previous methods. However, due to the lack of standard benchmarks for circuits and limited data availability for many circuits, it should be noted that this comparison should be viewed as a general guide rather than a comprehensive evaluation. Nonetheless, it can be observed that our method demonstrates superior performance in comparison to previous methods, while also maintaining a small data size.

For each circuit, we also provide the following additional information:
\begin{enumerate}
  \item The size of the train dataset, number of parameters, number of points per parameter in Table \ref{table_summary},
  \item Average performance error across all the circuits in Table  \ref{table_average_perf},
  \item Performance metric range for every circuit in Table  \ref{amp_r_table},\ref{nonl_r_table},
  \item Parameters range for every circuit in Table  \ref{Variables}.
\end{enumerate}

\begin{table*}[!hbt]
 \small
% \singlespacing
 \centering
\caption{Average performance error $\%$  for every circuit using $\bar{D}^{*}_{\epsilon}$ }\label{table_average_perf}
\begin{tabular}{|p{1.4cm}|p{1.8cm}|p{1.8cm}|p{1.8cm}|p{1.8cm}|p{1.8cm}|p{1.8cm}|p{1.8cm}|}
\hline
\textbf{Err/Circuit} & \textbf{CS} & \textbf{Cascode} &\textbf{Two Stage} & \textbf{LNA}  &\textbf{PA}   &\textbf{Mixer} &\textbf{VCO} \\
\hline
 Mean Error & 0.06 $ \pm$ 0.0 &  0.04 $ \pm$ 0.0 &  0.03 $ \pm$ 0.0 &  0.0 $ \pm$ 0.0 &  1.29 $ \pm$ 0.002 &  0.01 $ \pm$ 0.0 &  1.13 $ \pm$ 0.002 \\
\hline
\end{tabular}
\end{table*}

 \begin{center}
\begin{table}[h!]
\centering
\caption{Circuit complexity comparison}\label{table_summary}
\begin{tabular}{|c|c|c|c|c|c|c|c|}
\hline
  \textbf{Circuit} & \textbf{CS} & \textbf{Cascode} & \textbf{Two-Stage} & \textbf{LNA} & \textbf{PA} & \textbf{Mixer} & \textbf{VCO}\\
     \hline
    
        \textbf{train-data size}  & 3340 &  4080 & 616 & 4096 & 3528 & 3136 &4096  \\
      \hline
         \textbf{Number of Parameters}  & 2 & 3 & 3 & 4 & 4 & 3 & 4\\
      \hline
         \textbf{Points per parameter}  & 1670 & 1360 & 205.3 & 1024 & 882 & 1045.3 & 1024\\
      \hline
\end{tabular}
    \label{train-data}
\end{table}
\end{center}

\begin{table}[h!]
\centering
\caption{ Comparing Results with the previous work }\label{comparing_101}

\begin{tabular}{|c|c||c|c|c|}
\hline
   & \textbf{Performance Metric} & This Work (\%) & Best Reported (\%) & Related Works\\
 \hline
  \multirow{3}{*}{\rotatebox[origin=c]{90}{\textbf{CS}}} & Gain &  0.135$\pm$0.035 & \multirow{3}{*}{$<2.6$} &  \multirow{3}{*}{\citet{Devi2021}} \\
   \cline{2-3}
     & Bandwidth & 0.048 $\pm$ 0.12 &  & \\
      \cline{2-3}
      & Power Consumption & 0.0044$\pm$ 0.002 &  &  \\
      \cline{1-5}
  \multirow{3}{*}{\rotatebox[origin=c]{90}{\textbf{Cascode}}} & Gain & 0.0471$\pm$0.02 &\multirow{3}{*}{$\approx$ 1} & \multirow{3}{*}{\shortstack[c]{\citet{Lourenco2019}\\\citet{mina2022}}} \\
   \cline{2-3}
     & Bandwidth & 0.0433$\pm$0.01 &   & \\
      \cline{2-3}
      & Power Consumption & 0.0222$\pm$0.007 &  &  \\
      \cline{1-5}
  \multirow{3}{*}{\rotatebox[origin=c]{90}{\textbf{2-Stage}}} & Gain & 0.0226$\pm$0.018 & 1.1  & \citet{Fukuda2017} \\
   \cline{2-4}
     & Bandwidth & 0.00001 $\pm$ 0.000003 & NA &  \\
      \cline{2-4}
      & Power Consumption & 0.0716$\pm$0.031 & 3.7 & \citet{HarshaHarish}   \\
      \cline{1-5}
   \multirow{3}{*}{\rotatebox[origin=c]{90}{\textbf{LNA}}} & $G_{T}$ & 0.0028$\pm$0.001 & \multirow{3}{*}{ $<$ 5} &  \multirow{3}{*}{\citet{Dumesnil2014}} \\
   \cline{2-3}
     & $S_{11}$ & 0.007$\pm$0.001 &  & \\
      \cline{2-3}
      & NF & 0.002$\pm$0.0005 &  &  \\
      \cline{1-5}
   \multirow{3}{*}{\rotatebox[origin=c]{90}{\textbf{PA}}} & Power Gain & 1.207 $\pm$ 0.14
 &\multirow{3}{*}{NA}& \multirow{3}{*}{NA} \\
   \cline{2-3}
     & Drain Efficiency  & 1.361 $\pm$ 0.2 &  & \\
      \cline{2-3}
      & PAE & 1.30 $\pm $ 0.19 &  & \\
      \cline{1-5}  
 \multirow{3}{*}{\rotatebox[origin=c]{90}{\textbf{Mixer}}} & Conversion Gain & 0.005$\pm$0.0035 & \multirow{3}{*}{NA}   & \multirow{3}{*}{NA} \\
   \cline{2-3}
     & Power Consumption  & 0.86$\pm$0.002 & & \\
      \cline{2-3}
      & Swing  & 0.66$\pm$ 0.005 &  &  \\
      \cline{1-5}
 \multirow{3}{*}{\rotatebox[origin=c]{90}{\textbf{VCO}}} & Power Consumption  & 0.017$\pm$ 0.1423 & \multirow{3}{*}{NA}   & \multirow{3}{*}{NA} \\
   \cline{2-3}
     & Output Power & 0.1423$\pm$0.01 &  & \\
      \cline{2-3}
      & Tuning Range  & 3.226$\pm$0.6293 &  &    \\
      \cline{1-5}
\end{tabular}
    \label{previousworks}
\end{table}
\clearpage

\begin{table}[H]
\centering
\caption{ Range of Analog Voltage Amplifiers Performance Metrics}\label{range_1}
\resizebox{0.5\columnwidth}{!}{%
%\colorbox{purple!20}{
\begin{tabular}{|c||c|c|c|c|c|c|}
\hline
  \multirow{2}{*} {\textbf{Performance Metric}} & 
  \multicolumn{2}{c|}{\textbf{CS }} & 
  \multicolumn{2}{c|}{\textbf{Cascode}} &
  \multicolumn{2}{c|}{\textbf{Two-Stage }} \\
  \cline{2-7}
   
      &Min&Max&Min&Max&Min&Max \\
     \hline
     Gain (db) & 5.25 & 15.14 & 20.94 & 28.23 & 41.28 & 73.82\\
     \hline
      Bandwidth (Hz) & 83.7M & 5.99G & 2.17G & 8.5G  & 12.1M & 1.01G \\
      \hline
       Power Consumption (mW) & 0.57 & 1.34 & 0.38 & 0.56   & 1.32 & 2.00\\
       \hline
\end{tabular}
}
%}
    \label{amp_r_table}
\end{table}

%%%%%%%%%%%%%%%%%%%%%%%%%%%%%%%%%%%
% Amplifier Performance Metrics Ranges Table

\begin{table}[H]
\centering
\caption{Range of Non-linear Circuits Performance Metrics }\label{range_2}
\resizebox{0.5\columnwidth}{!}{%

\begin{tabular}{|c|c||c|c|}
\hline
   & \textbf{Performance Metric} & Min & Max\\
 
      \cline{1-4}
   \multirow{3}{*}{\rotatebox[origin=c]{90}{\textbf{LNA}}} & Power Gain(db) & 12.76 & 15.8  \\
   \cline{2-4}
     & $S_{11}$ & -19.1 & -17.3 \\
      \cline{2-4}
      & NF(db) & 2.154 & 2.39  \\
      \cline{1-4}

 \multirow{3}{*}{\rotatebox[origin=c]{90}{\textbf{PA}}} & Power Gain (db) & 5.165 & 18.65  \\
   \cline{2-4}
     & Drain Efficiency (\%) & 9.39 & 33.92 \\
      \cline{2-4}
      & PAE(\%) & 3.79 & 28.67  \\
      \cline{1-4}
 \multirow{3}{*}{\rotatebox[origin=c]{90}{\textbf{Mixer}}} & Conversion Gain & 0.61 & 5.95  \\
   \cline{2-4}
     & Power Consumption (mW) & 0.11 & 7.32 \\
      \cline{2-4}
      & Swing (mV) & 0.61 & 5.95  \\
      \cline{1-4}
 \multirow{3}{*}{\rotatebox[origin=c]{90}{\textbf{VCO}}} & Power Consumption (mW) & 3.9 & 12.3  \\
   \cline{2-4}
     & Output Power(mW) & 5.11 & 19.67 \\
      \cline{2-4}
      & Tuning Range (Hz) & 451K & 440M  \\
      \cline{1-4}
      
\end{tabular}

}
    \label{nonl_r_table}
\end{table}
%%%%%%%%%%%%%%%%%%%%%%%%%%%%%%%%%%%%%

%Design Parameters Table

\begin{table}[H]
  \centering
  \caption{ Design Parameters and  Range of Variations }\label{range_3}
  \resizebox{0.5\columnwidth}{!}{%
       \begin{tabular}{|c|c||c|c|c|}
                 \hline
    \textbf{Circuit} & \textbf{Variable} & Start & Step & End\\
    \hline
\multirow{2}{*}{\textbf{CS}} & $M_{1}$[w] & 2.8um & 0.2um & 6.6um \\
   \cline{2-5}
     & $R_{D}$ & 620$\Omega$ & 5$\Omega$  & 1450$\Omega$ \\
      \cline{1-5}
\multirow{3}{*}{\textbf{Cascode}} & $M_{1}$ [w] & 8um & 0.25um & 11.5um \\
   \cline{2-5}
     &  $M_{2}$ [w] & 2um & 0.2um & 5um\\
      \cline{2-5}
      & $R_{D}$ & 9k$\Omega$ & 125$\Omega$ & 11k$\Omega$ \\
      \cline{1-5}
\multirow{3}{*}{\textbf{Two-Stage }} & $M_{1}$[w] & 25um & 0.5um & 30um \\
   \cline{2-5}
     & $M_{2}$[w] & 52um & 0.5um & 55.5um\\
      \cline{2-5}
      & $M_{T}$[w] & 6um & 0.5um & 9um  \\
      \cline{1-5}
\multirow{4}{*}{\textbf{LNA}} & $M_{1,2}$[w] & 73um & 0.5um & 76.5um \\
   \cline{2-5}
     & $L_{g}$ & 9.4nH & 0.2nH & 10.8nH \\
      \cline{2-5}
       & $L_{s}$ & 747pH & 1pH & 754pH \\
      \cline{2-5}
       & $L_{d}$ & 3.7nH & 0.1nH & 4.4nH  \\
      \cline{1-5}
 \multirow{4}{*}{\textbf{PA}} & $M_{1,2,3,4}$[w] & 18um & 0.5um & 22um \\
   \cline{2-5}
     & $M_{5,6,7,8}$[w] & 27um & 1um & 34um \\
      \cline{2-5}
       & $V_{b1}$ & 785mV & 5mV & 815mV \\
      \cline{2-5}
      & $V_{b2}$ & 760mV & 5mV & 790mV  \\
      \cline{1-5}
      
 \multirow{3}{*}{\textbf{Mixer}} & $M_{1}$[w] & 8.55um & 0.45um & 11.7um  \\
   \cline{2-5}
    &$M_{T}$[w] & 17.1um & 0.9um & 23.4um\\
         \cline{2-5}
     & $V_{RF,DC}$ & 630mV & 30mV & 810mV \\
      \cline{2-5}
      & R  & 240 $\Omega$ & 40$\Omega$ & 520$\Omega$\\
      \cline{1-5}

 \multirow{4}{*}{\textbf{VCO}} & $M_{1}$[w] & 8.55um & 0.45um & 11.7um\\
   \cline{2-5}
    & $M_{T}$[w] & 145um & 2um & 159um \\
      \cline{2-5}
      &  $M_{V}$[w]& 73um & 2um & 87um  \\
       \cline{2-5}
        & L & 3.6nH & 0.1nH & 4.3nH \\
      \cline{1-5}
      
\end{tabular}
}
    \label{Variables}
\end{table}

%%%%%%%%%%%%%%%%%%%%%%%NMOS%%%%%%%%%%%%%%%%%%%%%%%

% \begin{figure}[!hbt]
%     \centering
%           \includegraphics[width=0.8\columnwidth]{fig/p2_methods/two_stage/subset-0.9-margin.png}%
%           \label{fig:left}%
%     \caption{TwoStage, comparing methods}
%     \label{p2_two_stage3}
% \end{figure}

%%%%%%%%%%%%%%%%%%%%%%%%%%%%%%%%%%%%%%%%%%%%%%%%%

%%%%%%%%%%%%%%%%%CASCODE%%%%%%%%%%%%%%%%%%%%%%%%%%%

%%%%%%%%%%%%%%%%%%%%%%%%%%%%%%%%%%%%%%%%%%%%%%%%%%%%%

%%%%%%%%%%%%%%%%%%%%%%TWO STAGE%%%%%%%%%%%%%%%%%%%%%%%%

%%%%%%%%%%%%%%%%%%%%%%%%%%%%%%%%%%%%%%%%%%%%%%%%%%%%%%%%%

%%%LNA%%%

%%%%%%

%%%PA%%%%

%%%%%

%%%MIXER%%%%

%%%%%

%VCO%%%

%%%

% \begin{equation}
% D^m_\epsilon = D_0   \bigcup_{i=1}^m D_\epsilon[u_i]; \quad u_i \sim U^k \text{ i.i.d}.
% \end{equation}

\end{document}